%% file: RSS_main_file.tex
\documentclass[conference]{IEEEtran}
\usepackage{times}

\usepackage[numbers]{natbib}
\usepackage{multicol}
\usepackage[bookmarks=true]{hyperref}
\usepackage{microtype}
\usepackage{graphicx}
\usepackage{subfigure}
\usepackage{booktabs} % for professional tables
\usepackage{bm}
\usepackage{hyperref}
\usepackage{amsmath}
\usepackage{mathtools}
\usepackage{xcolor}

\usepackage{algorithm}
\usepackage{algpseudocode}
\usepackage{centernot}

\begin{document}

\title{Fast and Feature-Complete Differentiable Physics for Articulated Rigid Bodies with Contact}

% You will get a Paper-ID when submitting a pdf file to the conference system
%\author{Author Names Omitted for Anonymous Review. Paper-ID 117}

% \author{\authorblockN{Keenon Werling}
% \authorblockA{Stanford University\\
% Email: keenon@stanford.edu}
% \and
% \authorblockN{Dalton Omens}
% \authorblockA{Stanford University\\
%  Email: domens@stanford.edu}
% \and
% \authorblockN{Jeongseok Lee}
% \authorblockA{Stanford\\

% avoiding spaces at the end of the author lines is not a problem with
% conference papers because we don't use \thanks or \IEEEmembership

\author{\authorblockN{Keenon Werling$^{1}$,
Dalton Omens$^{1}$,
Jeongseok Lee$^{2}$, 
Ioannis Exarchos$^{1}$ and
C. Karen Liu$^{1}$}
\authorblockA{$^{1}$Stanford University: \texttt{\{keenon, domens, exarchos, ckliu38\}@stanford.edu}}
\authorblockA{$^{2}$Robotics AI, Amazon: \texttt{jeoslee@amazon.com}}}
% \authorblockA{\authorrefmark{3}Starfleet Academy, San Francisco, California 96678-2391\\
% Telephone: (800) 555--1212, Fax: (888) 555--1212}
% \authorblockA{\authorrefmark{4}Tyrell Inc., 123 Replicant Street, Los Angeles, California 90210--4321}}

\maketitle

\input{defs}

\begin{abstract}
We present a fast and feature-complete differentiable physics engine, Nimble (\href{https://nimblephysics.org}{nimblephysics.org}), that supports Lagrangian dynamics and hard contact constraints for articulated rigid body simulation. Our differentiable physics engine offers a complete set of features that are typically only available in non-differentiable physics simulators commonly used by robotics applications. We solve contact constraints precisely using linear complementarity problems (LCPs). We present efficient and novel analytical gradients through the LCP formulation of inelastic contact that exploit the sparsity of the LCP solution. We support complex contact geometry, and gradients approximating continuous-time elastic collision. We also introduce a novel method to compute \emph{complementarity-aware gradients} that help downstream optimization tasks avoid stalling in saddle points. We show that an implementation of this combination in a fork of an existing physics engine (DART) is capable of a 87x single-core speedup over finite-differencing in computing analytical Jacobians for a single timestep, while preserving all the expressiveness of original DART.
\end{abstract}

\IEEEpeerreviewmaketitle

\input{introduction_v2}
\input{related_work}
\input{overview}
\input{LCP}

\input{collision_geometry}

\input{elastic_contact}

\input{evaluation}
\input{conclusion}

\section*{Acknowledgments}

%% Use plainnat to work nicely with natbib. 

\bibliographystyle{plainnat}
\bibliography{example_paper}
\newpage
\input{appendix}

\end{document}

\input{defs}
\twocolumn[
\icmltitle{Supplementary Material}



\icmlsetsymbol{equal}{*}

\begin{icmlauthorlist}
\icmlauthor{Keenon Werling}{st}
\icmlauthor{Ioannis Exarchos}{st}
\icmlauthor{Dalton Omens}{st}
\icmlauthor{C. Karen Liu}{st}
\end{icmlauthorlist}

\icmlaffiliation{st}{Computer Science Department, Stanford University, Stanford, California}

\icmlcorrespondingauthor{Keenon Werling}{keenon@cs.stanford.edu}
\icmlcorrespondingauthor{C. Karen Liu}{karenliu@cs.stanford.edu}

\icmlkeywords{Simulation, Robotics, Machine Learning}

\vskip 0.3in
]




\section{Derivatives of the Dynamics}

\begin{algorithm}[H]
	\caption{$\pd{}{q} \left( C \dot{q} + g \right)$}
	\label{alg:todo1}
    \For{$i = 1 \text{ to } n$}
        \State{$V_i = \AdTi V_{\lambda(i)} + S_i \dot{q}_i$}
        \State{$\dot{V}_i = \AdTi \dot{V}_{\lambda(i)} + S_i \dot{q}_i + \ad{V_i} S_i \dot{q}_i + \dot{S}_i \dot{q}_i$}
        \For{$j = 1 \text{ to } n$}
            \State{$\pd{V_i}{q_j} = \AdTi \pd{V_{\lambda(i)}}{q_j} - \ad{\pd{h_i}{q_j}} \AdTi V_{\lambda(i)}$}
            \State{$~~~~~~~~~ + \pd{S_i}{q_j} \dot{q}_i$}
            \State{$\pd{V_i}{\dot{q}_j} = \AdTi \pd{V_{\lambda(i)}}{\dot{q}_j} + S_i^k$}
            \State{$\pd{\dot{V}_i}{q_j} = \AdTi \pd{\dot{V}_{\lambda(i)}}{q_j} - \ad{\pd{h_i}{q_j}} \AdTi \dot{V}_{\lambda(i)}$}
            \State{$~~~~~~~~ + \ad{\pd{V_i}{q_j}} S_i \dot{q}_i + \ad{V_i} \pd{S_i}{q_j} \dot{q}_i + \pd{\dot{S}_i}{q_j} \dot{q}_i$}
            \State{$\pd{\dot{V}_i}{\dot{q}_j} = \AdTi \pd{\dot{V}_{\lambda(i)}}{\dot{q}_j}$}
            \State{$~~~~~~~~ + \ad{\pd{V_i}{\dot{q}_j}} S_i \dot{q}_i + \ad{V_i} S_i^k + \pd{\dot{S}_i}{\dot{q}_j} \dot{q}_i + \dot{S}_i^k$}
        \EndFor
    \EndFor
    \For{$i = n \text{ to } 1$}
        \State{$F_i = \I_i \dot{V}_i - \dad{V_i} \I_i V_i - F_i^g + \sum_{l \in \mu(i)} \dAd{T_{i,l}^{-1}} F_l$}
        \For{$j = 1 \text{ to } n$}
            \State{$\pd{F_i}{q_j} = \I_i \pd{\dot{V}_i}{q_j} - \dad{\pd{V_i}{q_j}} \I V_i - \dad{V_i} \I_i \pd{V_i}{q_j} - \pd{F^{\text{g}}}{q_j}$}
            \State{$~~~~~~~~~ + \sum_{l \in \mu(i)} \dAd{T_{i,l}^{-1}} \left( \pd{F_l}{q_j} - \dad{\pd{h_i}{q_j}} F_l \right)$}
            \State{$\pd{F_i}{\dot{q}_j} = \I_i \pd{\dot{V}_i}{\dot{q}_j} - \dad{\pd{V_i}{\dot{q}_j}} \I V_i - \dad{V_i} \I_i \pd{V_i}{\dot{q}_j}$}
            \State{$~~~~~~~~~ + \sum_{l \in \mu(i)} \dAd{T_{i,l}^{-1}} \pd{F_l}{q_j}$}
            \State{$\left[ \pd{\left( C \dot{q} + g \right)}{q} \right]_{i,j} = \pd{S_i}{q_j}^T F_i + S_i^T \pd{F_i}{q_j}$}
            \State{$\left[ \pd{\left( C \dot{q} + g \right)}{\dot{q}} \right]_{i,j} = S_i^T \pd{F_i}{\dot{q}_j}$}
        \EndFor
    \EndFor
\end{algorithm}

$\pd{h_i}{q_j} = \begin{cases} S_i^k & \text{if } q_j = q_i^k \\ 0, & \text{otherwise} \end{cases}$, $\pd{S_i}{q_j} = \pd{\dot{S}_i}{q_j} = 0$ when $q_j \centernot\in q_i$.

$\pd{\left( C \dot{q} \right)}{\tau} = 0$

\begin{algorithm}[H]
	\caption{$\pd{M}{q} \ddot{q}$}
	\label{alg:todo1}
    \For{$i = 1 \text{ to } n$}
        \State{$V_i = 0$}
        \State{$\dot{V}_i = \AdTi \dot{V}_{\lambda(i)} + S_i \ddot{q}_i$}
        \For{$j = 1 \text{ to } n$}
            \State{$\pd{V_i}{q_j} = 0$}
            \State{$\pd{\dot{V}_i}{q_j} = \AdTi \pd{\dot{V}_{\lambda(i)}}{q_j} - \ad{\pd{h_i}{q_j}} \AdTi \dot{V}_{\lambda(i)}$}
            \State{$~~~~~~~~ + \pd{S_i}{q_j} \ddot{q_i}$}
        \EndFor
    \EndFor
    \For{$i = n \text{ to } 1$}
        \State{$F_i = \I_i \dot{V}_i + \sum_{l \in \mu(i)} \dAd{T_{i,l}^{-1}} F_l$}
        \For{$j = 1 \text{ to } n$}
            \State{$\pd{F_i}{q_j} = \I_i \pd{\dot{V}_i}{q_j} - \pd{F^{\text{g}}}{q_j}$}
            \State{$~~~~~~~~~ + \sum_{l \in \mu(i)} \dAd{T_{i,l}^{-1}} \left( \pd{F_l}{q_j} - \dad{\pd{h_i}{q_j}} F_l \right)$}
            \State{$\left[ \pd{M}{q} \right]_{i,j} = \pd{S_i}{q_j}^T F_i + S_i^T \pd{F_i}{q_j}$}
        \EndFor
    \EndFor
\end{algorithm}

\begin{algorithm}[H]
	\caption{$\pd{M^{-1}}{q} x$}
	\label{alg:todo1}
    % \For{$i = 1 \text{ to } n$}
    %     \State{$V_i = \AdTi V_{\lambda(i)} + S_i \dot{q}_i$}
    %     % \For{$j = 1 \text{ to } n$}
    %     %     \State{$\pd{V_i}{q_j} = \AdTi \pd{V_{\lambda(i)}}{q_j} - \ad{\pd{h_i}{q_j}} \AdTi V_{\lambda(i)}$}
    %     % \EndFor
    % \EndFor
    % \State{Initialization: $\pd{\AB_n}{q_j} = 0$}
    \For{$i = n \text{ to } 1$}
        \State{$\AI_i = \I_i + \sum_{l \in \mu(i)} \dAd{T_{i,l}^{-1}} \Pi_l \Ad{T_{T_{i,l}}^{-1}}$}
        \State{$\AB_i = \sum_{l \in \mu(i)} \dAd{T_{i,l}^{-1}} \beta_l$}
        \State{$\Psi_i = \left( S_i^T \AI_i S_i \right)^{-1}$}
        \State{$\Pi_i = \AI_i - \AI_i S_i \Psi_i S_i^T \AI_i$}
        \State{$\alpha_i = x_i - S_i^T \AB_i$}
        \State{$\beta_i = \AB_i + \AI_i S_i \Psi_i \alpha_i$}
        \For{$j = 1 \text{ to } n$}
            \State{$\pd{\AI_i}{q_j} = \sum_{l \in \mu(i)} \dAd{T_{i,l}^{-1}} \left\{ \pd{\Pi_l}{q_j} - \Pi_l \ad{\pd{h_l}{q_j}} - \left( \Pi_l \ad{\pd{h_l}{q_j}} \right)^T \right\} \Ad{T_{i,l}^{-1}}$}
            \State{$\pd{\AB_i}{q_j} = \sum_{l \in \mu(i)} \dAd{T_{i,l}^{-1}} \left( \pd{\beta_l}{q_j} - \dad{\pd{h_l}{q_j}} \beta_l \right)$}
            \State{$\pd{\Psi_i}{q_j} = -\Psi_i \left\{ S_i^T \pd{\AI_i}{q_j} S_i + \pd{S_i}{q_j}^T \AI_i S_i + \left( \pd{S_i}{q_j}^T \AI_i S_i \right)^T \right\} \Psi_i$}
            \State{$\pd{\Pi_i}{q_j} = \pd{\AI_i}{q_j} - \AI_i S_i \pd{\Psi_i}{q_j} S_i^T \AI_i - \pd{\AI_i}{q_j} S_i \Psi_i S_i^T \AI_i$}
            \State{$~~~~~~ - \left( \pd{\AI_i}{q_j} S_i \Psi_i S_i^T \AI_i \right)^T - \AI_i \pd{S_i}{q_j} \Psi_i S_i^T \AI_i$}
            \State{$~~~~~~ - \left( \AI_i \pd{S_i}{q_j} \Psi_i S_i^T \AI_i \right)^T$}
            % \State{$\pd{\alpha_i}{q_j} = \pd{x_i}{q_j} - \pd{S_i}{q_j}^T \AB_i - S_i^T \pd{\AB_i}{q_j}$}
            \State{$\pd{\alpha_i}{q_j} = - \pd{S_i}{q_j}^T \AB_i - S_i^T \pd{\AB_i}{q_j}$}
            \State{$\pd{\beta_i}{q_j} = \pd{\AB_i}{q_j} + \pd{\AI_i}{q_j} S_i \Psi_i \alpha_i + \AI_i \pd{S_i}{q_j} \Psi_i \alpha_i$}
            \State{$~~~~~~ + \AI_i S_i \pd{\Psi_i}{q_j} \alpha_i + \AI_i S_i \Psi_i \pd{\alpha_i}{q_j}$}
        \EndFor
    \EndFor
    \For{$i = 1 \text{ to } n$}
        \State{$\ddot{q_i} = \Psi_i \left( \alpha_i - S_i^T \AI_i \AdTi \dot{V}_{\lambda(i)} \right)$}
        \State{$\dot{V}_i = \AdTi \dot{V}_{\lambda(i)} + S_i \ddot{q}_i $}
        \For{$j = 1 \text{ to } n$}
            \State{$\left[ \pd{M^{-1}}{q} x \right]_{i,j} = \pd{\Psi_i}{q_j} \left( \alpha_i - S_i^T \AI_i \AdTi \dot{V}_{\lambda(i)} \right)$}
            \State{$~~~~~~ + \Psi_i \left\{ \pd{\alpha_i}{q_j} - \left( \pd{S_i}{q_j}^T \AI_i + S_i^T \pd{\AI_i}{q_j} \right) \AdTi \dot{V}_{\lambda(i)} \right.$}
            \State{$~~~~~~ \left. - S_i^T \AI_i \left( \AdTi \pd{\dot{V}_{\lambda(i)}}{q_j} - \ad{\pd{h_i}{q_j}} \AdTi \dot{V}_{\lambda(i)} \right) \right\}$}
            \State{$\pd{\dot{V}_i}{q_j} = \AdTi \pd{\dot{V}_{\lambda(i)}}{q_j} - \ad{\pd{h_i}{q_j}} \AdTi \dot{V}_{\lambda(i)}$}
            \State{$~~~~~~ + \pd{S_i}{q_j} \ddot{q} + S_i^T \left[ \pd{M^{-1}}{q} x \right]_{i,j}$}
        \EndFor
    \EndFor
\end{algorithm}

$\pd{M^{-1}}{\dot{q}} = \pd{M^{-1} }{\tau}  = 0$.


\bibliography{example_paper}
\bibliographystyle{icml2019}

%% file: defs.tex
%\newcommand{\cmt}[1]{\textcolor{red}{\textbf {#1}}}
%\usepackage[usenames] {color}
%\definecolor{purple}{rgb}{0.4,0.2,0.8}
\definecolor{orange}{rgb}{0.9, 0.3, 0.0}
\definecolor{darkblue}{rgb}{0.0, 0.0, 0.53}
\definecolor{darkgreen}{rgb}{0.0, 0.53, 0.0}
\newcommand{\old}[1]{\textcolor{red}{\textbf {\st{#1}}}}
\newcommand{\new}[1]{\textcolor{red}{#1}}
\newcommand{\note}[1]{\cmt{Note: #1}}
\newcommand{\karen}[1]{\textcolor{red}{{[Karen: #1]}}}
\newcommand{\keenon}[1]{\textcolor{orange}{{[Keenon: #1]}}}
\newcommand{\js}[1]{\textcolor{blue}{{[JS: #1]}}}
\newcommand{\dalton}[1]{\textcolor{darkgreen}{{[Dalton: #1]}}}
\newcommand{\amy}[1]{\textcolor{purple}{[Amy: #1]}}
\newcommand{\eqnref}[1]{Equation~(\ref{eqn:#1})}
%% ignore text
\long\def\ignorethis#1{}

\newcommand{\sect}[1]{Section~\ref{#1}}
\newcommand{\myparagraph}[1]{\vspace{5pt}\noindent\textbf{#1}}
%% abbreviations
\makeatletter
\DeclareRobustCommand\onedot{\futurelet\@let@token\@onedot}
\def\@onedot{\ifx\@let@token.\else.\null\fi\xspace}

\def\iid{i.i.d\onedot}
\def\eg{e.g\onedot} \def\Eg{E.g\onedot}
\def\ie{i.e\onedot} \def\Ie{I.e\onedot}
\def\cf{\emph{c.f}\onedot} \def\Cf{\emph{C.f}\onedot}
\def\etc{etc\onedot} \def\vs{vs\onedot}
\def\wrt{w.r.t\onedot} \def\dof{d.o.f\onedot}
\def\etal{et al\onedot}
\makeatother

%% reference shortcuts
\newcommand{\figtodo}[1]{\framebox[0.8\columnwidth]{\rule{0pt}{1in}#1}}
\newcommand{\figref}[1]{Figure~\ref{fig:#1}}
\newcommand{\secref}[1]{Section~\ref{sec:#1}}

%% frequently used mathematical structures
% \newcommand{\vc}[1]{\ensuremath{\mathbf{#1}}}
\newcommand{\vc}[1]{\ensuremath{\boldsymbol{#1}}}
\newcommand{\pd}[2]{\ensuremath{\frac{\partial{#1}}{\partial{#2}}}}
\newcommand{\pdd}[3]{\ensuremath{\frac{\partial^2{#1}}{\partial{#2}\,\partial{#3}}}}

% math macros
\newcommand{\vEndEff}{\ensuremath{\vc{d}}}
\newcommand{\vRelMove}{\ensuremath{\vc{r}}}
\newcommand{\sSet}{\ensuremath{S}}

\newcommand{\vControl}{\ensuremath{\vc{u}}}
\newcommand{\vPoint}{\ensuremath{\vc{p}}}
\newcommand{\sSpringCoef}{{\ensuremath{k_{s}}}}
\newcommand{\sDamperCoef}{{\ensuremath{k_{d}}}}
\newcommand{\vHandle}{\ensuremath{\vc{h}}}
\newcommand{\vForce}{\ensuremath{\vc{f}}}

\newcommand{\mTransChain}{\ensuremath{\vc{W}}}
\newcommand{\mRotateTrans}{\ensuremath{\vc{R}}}
\newcommand{\sJoint}{\ensuremath{q}}
\newcommand{\vJoint}{\ensuremath{\vc{q}}}
\newcommand{\mJoint}{\ensuremath{\vc{Q}}}
\newcommand{\mMass}{\ensuremath{\vc{M}}}
\newcommand{\sMass}{\ensuremath{{m}}}
\newcommand{\vGravity}{\ensuremath{\vc{g}}}
\newcommand{\vConstr}{\ensuremath{\vc{C}}}
\newcommand{\sConstr}{\ensuremath{C}}
\newcommand{\vCOM}{\ensuremath{\vc{x}}}
\newcommand{\sGeneralForce}[1]{\ensuremath{Q_{#1}}}
\newcommand{\vStateVar}{\ensuremath{\vc{y}}}
\newcommand{\vControlVar}{\ensuremath{\vc{u}}}
\newcommand{\argmax}{\operatornamewithlimits{argmax}}
\newcommand{\argmin}{\operatornamewithlimits{argmin}}

\newcommand{\tr}[1]{\ensuremath{\mathrm{tr}\left(#1\right)}}

%%%%%%%%%%%%%%%%%%%%%%%%%%%%%%%%%%%%%%%%%%%%%%%%%%%%%%%%%%%%%%%%%%%
%
% Macros for Lie group dynamics formulation
%
%%%%%%%%%%%%%%%%%%%%%%%%%%%%%%%%%%%%%%%%%%%%%%%%%%%%%%%%%%%%%%%%%%%

\newcommand{\Ad}[1]{\text{Ad}_{#1}}
\newcommand{\AdTi}{\text{Ad}_{\vc{T}_{\lambda(i),i}^{-1}}}

\newcommand{\dAd}[1]{\text{Ad}_{#1}^{*}}
\newcommand{\dAdTi}{\text{Ad}_{T_{\lambda(i),i}^{-1}}^{*}}

\newcommand{\ad}[1]{\text{ad}_{#1}}
\newcommand{\dad}[1]{\text{ad}_{#1}^{*}}

\newcommand{\I}{\vc{\mathcal{I}}}  % spatial inertia (6x6)
\newcommand{\AI}{\hat{\vc{\mathcal{I}}}} % articulated spatial inertia (6x6)
\newcommand{\AB}{\hat{\vc{\mathcal{B}}}} % articulated biased force (6x1)

%%%%%%%%%%%%%%%%%%%%%%%%%%%%%%%%%%%%%%%%%%%%%%%%%%%%%%%%%%%%%%%%%%%
%
% Here are a bunch of macros, mostly for math.
%
%%%%%%%%%%%%%%%%%%%%%%%%%%%%%%%%%%%%%%%%%%%%%%%%%%%%%%%%%%%%%%%%%%%

\renewcommand{\choose}[2]{\ensuremath{\left(\begin{array}{c} #1 \\ #2 \end{array} \right )}}

\newcommand{\gauss}[3]{\ensuremath{\mathcal{N}(#1 | #2 ; #3)}}

\newcommand{\pctab}{\hspace{0.2in}}
\newenvironment{pseudocode} {\begin{center} \begin{minipage}{\textwidth}
                             \normalsize \vspace{-2\baselineskip} \begin{tabbing}
                             \pctab \= \pctab \= \pctab \= \pctab \=
                             \pctab \= \pctab \= \pctab \= \pctab \= \\}
                            {\end{tabbing} \vspace{-2\baselineskip}
                             \end{minipage} \end{center}}
\newenvironment{items}      {\begin{list}{$\bullet$}
                              {\setlength{\partopsep}{\parskip}
                                \setlength{\parsep}{\parskip}
                                \setlength{\topsep}{0pt}
                                \setlength{\itemsep}{0pt}
                                \settowidth{\labelwidth}{$\bullet$}
                                \setlength{\labelsep}{1ex}
                                \setlength{\leftmargin}{\labelwidth}
                                \addtolength{\leftmargin}{\labelsep}
                                }
                              }
                            {\end{list}}
\newcommand{\newfun}[3]{\noindent\vspace{0pt}\fbox{\begin{minipage}{3.3truein}\vspace{#1}~ {#3}~\vspace{12pt}\end{minipage}}\vspace{#2}}

\newcommand{\norm}[1]{\left\lVert#1\right\rVert}

\newcommand{\key}{\textbf}
\newcommand{\fun}{\textsc}

%\def\shortcite{\def\citename##1{}\@internalcite}

% Local Variables:
% TeX-master: "paper"
% End:

%% file: introduction_v2.tex
\section{Introduction}
\label{intro}

Many modern robotics problems are optimization problems. Finding optimal trajectories, guessing the physical parameters of a world that best fits our observed data, or designing a control policy to optimally respond to a dynamic environment are all types of optimization problems. Optimization methods can be sorted into two buckets: gradient-based, and gradient-free. Despite the well-known drawbacks of gradient-free methods (high sample complexity and noisy solutions), they remain popular in robotics because physics engines with the necessary features to model complex robots are generally non-differentiable. When gradients are required, we typically approximate them using finite differencing \cite{todorov2012mujoco}.

Recent years have seen many differentiable physics engines published \cite{degrave2019differentiable, hu2019taichi, heiden2020augmenting, toussaint2018differentiable, song2020learning, de2018end, qiao2020scalable, millard2020automatic, geilinger2020add}, but none has yet gained traction as a replacement for popular non-differentiable engines \cite{bullet,todorov2012mujoco,dart}. We hypothesize that an ideal differentiable physics engine needs to implement an equivalent feature set to existing popular non-differentiable engines, as well as provide excellent computational efficiency, in order to gain adoption.
In this paper, we extend the existing physics engine DART \cite{dart}, which is commonly used in robotics and graphics communities, and make it differentiable. Our resulting engine supports all features available to the forward simulation process, meaning existing code and applications will remain compatible while also enjoying new capabilities enabled by efficient analytical differentiability.

A fully-featured physics engine like DART is complex and has many components that must be differentiated. Some components (such as collision detection and contact force computation) are not naively differentiable, but we show that under very reasonable assumptions we can compute useful Jacobians regardless. 
In order to differentiate through contacts and collision forces, we introduce an efficient method for differentiating the contact Linear Complementarity Problem (LCP) that exploits sparsity, as well as novel contact geometry algorithms and an efficient continuous-time approximation for elastic collisions. As a result, our engine is able to compute gradients with hard contact constraints up to 87 times faster than finite differencing methods, depending on the size of the dynamic system. Our relative speedup over finite differencing grows as the system complexity grows.
In deriving our novel method to differentiate through the LCP, we also gain an understanding of the nature of contact dynamics Jacobians that allows us to propose heuristic “complementarity-aware gradients” that may be good search directions to try, if a downstream optimizer is getting stuck in a saddle point.

We provide an open-source implementation of all of these ideas, as well as derivations from previous work \cite{kim2012lie, carpentier2018analytical}, in a fully differentiable fork of the DART physics engine, which we call Nimble. Code and documentation are available at \href{https://nimblephysics.org}{nimblephysics.org}.

\begin{figure}
\centering
\includegraphics[width=8.2cm]{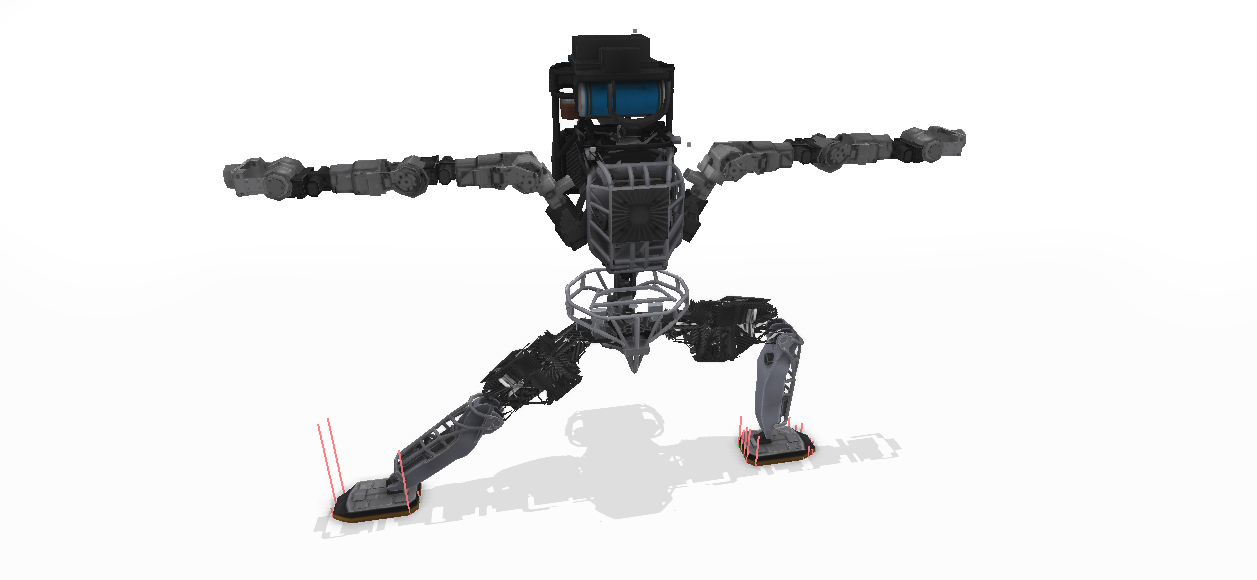}
\vspace{-0.2cm}
\caption{Boston Dynamics' Atlas Robot learning to do yoga, being simulated by our engine. This robot has 34 mesh colliders and 32 degrees of freedom. This freeze frame contains 24 contact points, 12 per foot. Even with all that complexity, we are able to compute the Jacobians of dynamics on this robot \textbf{87x faster} on a single CPU core using the analytical methods introduced in this paper than by finite differencing (only \textbf{8.5ms} vs 749ms for finite differencing).}
\vspace{-0.2cm}
\label{fig:simstep}
\end{figure}

To summarize, our contributions are as follows:
\begin{itemize}
    \item A novel and fast method for local differentiability of LCPs that exploits the sparsity of the LCP solution, which gives us efficient gradients through static and sliding contacts and friction without changing traditional forward-simulation formulations. \textit{Section \ref{sec:lcp}}
    \item A novel method that manipulates gradient computation to help downstream optimization problems escape saddle points due to discrete contact states. \textit{Section \ref{sec:complementarity-aware}}
    \item Fast geometric analytical gradients through collision detection algorithms which support various types of 3D geometry and meshes. \textit{Section \ref{sec:collision}}
    \item A novel analytical approximation of continuous-time gradients through elastic contact, which otherwise can lead to errors in discrete-time systems. \textit{Section \ref{sec:bounces}}
    \item An open-source implementation of all of our proposed methods (along with analytical gradients through Featherstone first described in GEAR \cite{kim2012lie}) in a fork of the DART physics engine which we call Nimble. We have created Python bindings and a PyPI package, \texttt{pip install nimblephysics}, for ease of use.
\end{itemize}

%% file: related_work.tex
\section{Related Work}
\begin{table}[t]
\label{tab:engine-comparison}
\vskip 0.15in
\begin{center}
\begin{small}
\begin{tabular}{lcccc}
\toprule
Engine & Contact & Dynamic & Collision & Gradients \\
& Force & State & Geometry & Method \\
\midrule
\midrule
MuJoCo  & customized & \textbf{generalized} & \textbf{complete} & finite\\
Degrave & impulse & cartesian & primitives & auto \\
DiffTaichi  & impulse & caresian & primitives & auto \\
Heiden & iter LCP & \textbf{generalized} & primitives & auto \\
de A. B.-P. & \textbf{direct LCP} & cartesian & primitives & \textbf{symbolic} \\
Geilinger & customized & \textbf{generalized} & primitives & \textbf{symbolic} \\
\midrule
Ours & \textbf{direct LCP} & \textbf{generalized} & \textbf{complete} & \textbf{symbolic}  \\
\bottomrule
\end{tabular}
\end{small}
\end{center}
\caption{Differentiable engines supporting articulated rigid bodies}
\vskip -0.1in
\end{table}

% \begin{table*}[t]
% \label{engine-comparison}
% \vskip 0.15in
% \begin{center}
% \begin{small}
% \begin{tabular}{lcccc}
% \toprule
% Engine & Non-ground & Featherstone & Solver & Analytical Gradients \\
% \midrule
% A.D.D. \cite{geilinger2020add}    & $\times$ & $\surd$ & Hybrid & $\surd$ \\
% Tiny \cite{heiden2020augmenting} & $\surd$ & $\times$ & Impulse & $\surd$ \\
% DiffTaichi \cite{difftaichi} & $\times$ & $\times$ & Impulse & $\surd$ \\
% ``Theano Based'' \cite{degrave2019differentiable} & $\times$ & $\times$ & Impulse & $\surd$ \\
% ``Optnet Based'' \cite{de2018end} & $\surd$ & $\times$ & LCP & $\surd$ \\
% GEAR \cite{kim2012lie} & N/A & $\surd$ & N/A & $\surd$ \\
% MuJoCo \cite{todorov2012mujoco} & $\surd$ & $\surd$ & Hybrid & $\times$ \\
% \bottomrule
% \end{tabular}
% \end{small}
% \end{center}
% \caption{Trade off comparison of published differentiable physics engines}
% \vskip -0.1in
% \end{table*}
%\karen{JS, can you say something about Carpentier's work I asked you about?}
%\js{Sure. I wanted to verify their algorithm before adding something here. Today, I confirmed that their algorithm is equivalent in terms of the result, and it's 2.x times faster, for a 10-DOF serial chain, than the full derivative of $M^{-1}/q$ using the forward dynamics algorithm. Let me add something here tomorrow.}
Differentiable physics simulation has been investigated previously in many different fields, including mechanical engineering \cite{hermans2014automated}, robotics \cite{heiden2019interactive}, physics \cite{jarny1991general,iollo2001aerodynamic} and computer graphics \cite{popovic2000interactive,mcnamara2004fluid}. Enabled by recent advances in automatic differentiation methods and libraries \cite{paszke2017automatic, al2016theano, difftaichi}, a number of differentiable physics engines have been proposed to solve control and parameter estimation problems for rigid bodies \cite{kim2012lie,degrave2019differentiable,difftaichi,heiden2020augmenting,de2018end,geilinger2020add,qiao2020scalable} and non-rigid bodies \cite{schenck2018spnets,liang2019,hahn2019real2sim,hu2019chainqueen,qiao2020scalable,holl2020learning,geilinger2020add}. While they share a similar high-level goal of solving ``inverse problems'', the features and functionality provided by these engines vary widely, including the variations in contact handling, state space parameterization and collision geometry support. Table \ref{tab:engine-comparison} highlights the differences in a few differentiable physics engines that have demonstrated the ability to simulate articulated rigid bodies with contact. Based on the functionalities each engine intends to support, the approaches to computing gradients can be organized in following categories.

\textbf{Finite-differencing} is a straightforward way to approximate gradients of a function. For a feature-complete physics engine, where analytical gradients are complex to obtain, finite-differencing provides a simpler method. For example, a widely used physics engine, MuJoCo \cite{todorov2012mujoco}, supports gradient computation via finite differencing. However, finite-differencing tends to introduce round-off errors and performs poorly for a large number of input variables. 

\textbf{Automatic differentiation (auto-diff)} is a method for computing gradients of a sequence of elementary arithmetic operations or functions automatically. However, the constraint satisfaction problems required by many existing, feature-complete robotic physics engines are not supported by auto-diff libraries. To avoid this issue, many recent differentiable physics engines instead implement impulse-based contact handling, which could lead to numerical instability and constraint violation if the contact parameters are not tuned properly for the specific dynamic system and the simulation task. Degrave et al. \cite{degrave2019differentiable} implemented a rigid body simulator in the Theano framework \cite{al2016theano}, while DiffTaichi \cite{difftaichi} implemented a number of differentiable physics engines, including rigid bodies, extending the Taichi programming language \cite{hu2019taichi}, both representing dynamic equations in Cartesian coordinates and handling contact with impulse-based methods. In contrast, Tiny Differentiable Simulator \cite{heiden2020augmenting} models contacts as an LCP, but they solve the LCP iteratively via Projected Gauss Siedel (PGS) method \cite{jourdan1998gauss}, instead of directly solving a constraint satisfaction problem, making it possible to compute gradient using auto-diff libraries.

\textbf{Symbolic differentiation} is another way to compute gradients by directly differentiate mathematical expressions. For complex programs like Lagrangian dynamics with constraints formulated as a Differential Algebraic Equations, symbolic differentiation can be exceedingly difficult. Earlier work computed symbolic gradients for smooth dynamic systems \cite{kim2012lie}, and  \cite{carpentier2018analytical} simplified the computation of the derivative of the forward dynamics exploiting the derivative of the inverse dynamics. Symbolic differentiation becomes manageable when the gradients are only required within smooth contact modes \cite{toussaint2018differentiable} or a specific contact mode is assumed \cite{song2020learning}. Recently, Amos and Kolter proposed a method, Opt-Net, that back-propagates through the solution of an optimization problem to its input parameters \cite{amos2017optnet}. Building on Opt-Net, de Avila Belbute-Peres et al. \cite{de2018end} derived analytical gradients through LCP formulated as a QP. Their method enables differentiability for rigid body simulation with hard constraints, but their implementation represents 2D rigid bodies in Cartesian coordinates and only supports collisions with a plane, insufficient for simulating complex articulated rigid body systems. More importantly, computing gradients via QP requires solving a number of linear systems which does not take advantage of the sparsity of the LCP structure. Qiao et al. \cite{qiao2020scalable} built on \cite{amos2017optnet} and improved the performance of contact handling by breaking a large scene into smaller impact zones. A QP is solved for each impact zone to ensure that the geometry is not interpenetrating, but contact dynamics and conservation laws are not considered. Solving contacts for localized zones has been previously implemented in many existing physics engines \cite{todorov2012mujoco,bullet,dart}. Adapting the collision handling routine in DART, our method by default utilizes the localized contact zones to speed up the performance. Adjoint sensitivity analysis \cite{rackauckas2018comparison} has also been used for computing gradients of dynamics. Millard et al. \cite{millard2020automatic} combined auto-diff with adjoint sensitivity analysis to achieve faster gradient computation for higher-dof systems, but their method did not handle contact and collision. Geilinger et al. \cite{geilinger2020add} analytically computed derivatives through adjoint sensitivity analysis and proposed a differentiable physics engine with implicit forward integration and a customized frictional contact model that is natively differentiable.

\textbf{Approximating physics with neural networks} is a different approach towards differentiable physics engine. Instead of forward simulating a dynamic system from the first principles of Newtonian mechanics, a neural network is learned from training data. Examples of this approach include Battaglia et. al \cite{battaglia2016interaction}, Chang et. al. \cite{chang2016compositional}, and Mrowca et. al \cite{mrowca2018flexible}.

\textbf{Our engine} employs symbolic differentiation to compute gradients through every part of the engine using hand-tuned C++ code.
We introduce a novel method to differentiate the LCP analytically that takes advantage of the sparsity of the solution and is compatible with using direct methods to solve the LCP. 
% This reduces the big $\mathcal{O}$ runtime of our LCP forward/backward computation by a factor of $\mathcal{O}(n)$ \karen{I don't understand this part. What is $n$?}over methods that attempt to use jets to automatically differentiate an iterative solution to the LCP, such as Tiny Differentiable Simulation \cite{heiden2020augmenting}.
% v2
% Like Tiny Differentiable Simulator, its forward simulation integrates Lagrangian dynamics in generalized coordinates and solves for constraint forces via a LCP.
% However, Tiny Differentiable Simulator uses jets to automatically differentiate through an iterative PGS solver, which expands the computation required to solve the LCP by a multiple of $\mathcal{O}(n)$, where $n$ is the number of variables you wish to construct Jacobians against. By contrast, in this paper we introduce a novel method to differentiate the LCP analytically, which enables us to use direct methods to solve the LCP, and do so using natural numbers, saving a factor of $\mathcal{O}(n)$ on one of the most expensive parts of forward physics.
% v1
% However, our engine directly solves the LCP rather than employing an iterative method such as PGS, which convergence is sensitive to the simulation tasks, thereby requiring careful tuning of optimization parameters for each task.
In addition, our engine supports a richer set of geometry for collision and contact handling than has been previously available, including mesh-mesh and mesh-primitive collisions, in order to achieve a fully functional differentiable version of the DART physics engine for robotic applications.

%% file: overview.tex
\section{Overview}

%A typical forward simulation involves Collision Detection, Contact Handling, Forward Dynamics and Integration in each discretized timestep. Non-differentiability arises at Collision Detection and Contact Handling.

A physics engine can be thought of as a simple function that takes the current position $\vc{q}_t$, velocity $\dot{\vc{q}}_t$, control forces $\vc{\tau}$ and inertial properties $\vc{\mu}$, and returns the position and velocity at the next timestep, $\vc{q}_{t+1}$ and $\dot{\vc{q}}_{t+1}$:
\begin{equation}
P(\vc{q}_t, \dot{\vc{q}}_t, \vc{\tau}, \vc{\mu}) = [\vc{q}_{t+1}, \dot{\vc{q}}_{t+1}].
\end{equation}

In an engine with simple explicit time integration, our next position $\vc{q}_{t+1}$ is a trivial function of current position and velocity, $\vc{q}_{t+1} = \vc{q}_t + \Delta t \dot{\vc{q}}_{t}$ , where $\Delta t$ is the descritized time interval.

The computational work of the physics engine comes from solving for our next velocity, $\dot{\vc{q}}_{t+1}$. We are representing our articulated rigid body system in generalized coordinates using the following Lagrangian dynamic equation:
\begin{equation}
\label{eqn:EOM}
\begin{split}
\bm{M}(\bm{q}_t, \vc{\mu})\dot{\bm{q}}_{t+1} &= \bm{M}(\bm{q}_t, \vc{\mu})\dot{\bm{q}}_t - \Delta t(\bm{c}(\bm{q}_t, \dot{\bm{q}}_t, \vc{\mu}) - \bm{\tau}) \\
&+ \bm{J}^T(\bm{q}_t) \bm{f},
\end{split}
\end{equation}
where $\bm{M}$ is the mass matrix, $\bm{c}$ is the Coriolis and gravitational force, and $\bm{f}$ is the contact impulse transformed into the generalized coordinates by the contact Jacobian matrix $\bm{J}$. Note that multiple contact points and/or other constraint impulses can be trivially added to Equation \ref{eqn:EOM}.

Every term in Equation \ref{eqn:EOM} can be evaluated given $\vc{q}_t$, $\dot{\vc{q}}_t$ and $\bm{\tau}$ except for the contact impulse $\vc{f}$, which requires the engine to form and solve an LCP:
\begin{align}
    &\mathrm{find\;\;\;} \bm{f}, \bm{v}_{t+1} \nonumber \\
    &\mathrm{such\;that\;\;\;} \bm{f} \geq \bm{0}, \; \bm{v}_{t+1} \geq \bm{0}, \; \bm{f}^T\bm{v}_{t+1} = 0.
\end{align}

The velocity of a contact point at the next time step, $\bm{v}_{t+1}$, can be expressed as a linear function in $\bm{f}$
\begin{align}
    \bm{v}_{t+1} &= \bm{J}\dot{\bm{q}}_{t+1} = \bm{J}\bm{M}^{-1} \left( \bm{M}\dot{\bm{q}}_t - \Delta t (\bm{c} - \bm{\tau}) + \bm{J}^T\bm{f} \right) \nonumber \\
    &= \bm{A}\bm{f} + \bm{b},
\end{align}
where $\bm{A} = \bm{J}\bm{M}^{-1}\bm{J}^T$ and $\bm{b} = \bm{J}(\dot{\bm{q}}_t + \Delta t \bm{M}^{-1}(\bm{\tau} - \bm{c}))$. The LCP procedure can then be expressed as a function that maps $(\bm{A}, \bm{b})$ to the contact impulse $\bm{f}$:
\begin{equation}
\label{eqn:LCP}
    f_{\text{LCP}}\big(\bm{A}(\bm{q}_t, \vc{\mu}), \bm{b}(\vc{q}_t, \dot{\vc{q}}_t, \vc{\tau}, \vc{\mu})\big) = \bm{f}
\end{equation}

As such, the process of forward stepping is to find $\bm{f}$ and resulting $\dot{\bm{q}}_{t+1}$ that satisfy Equation \ref{eqn:EOM} and Equation \ref{eqn:LCP}.

\begin{figure}[ht]
\begin{center}
\centerline{\includegraphics[width=\columnwidth]{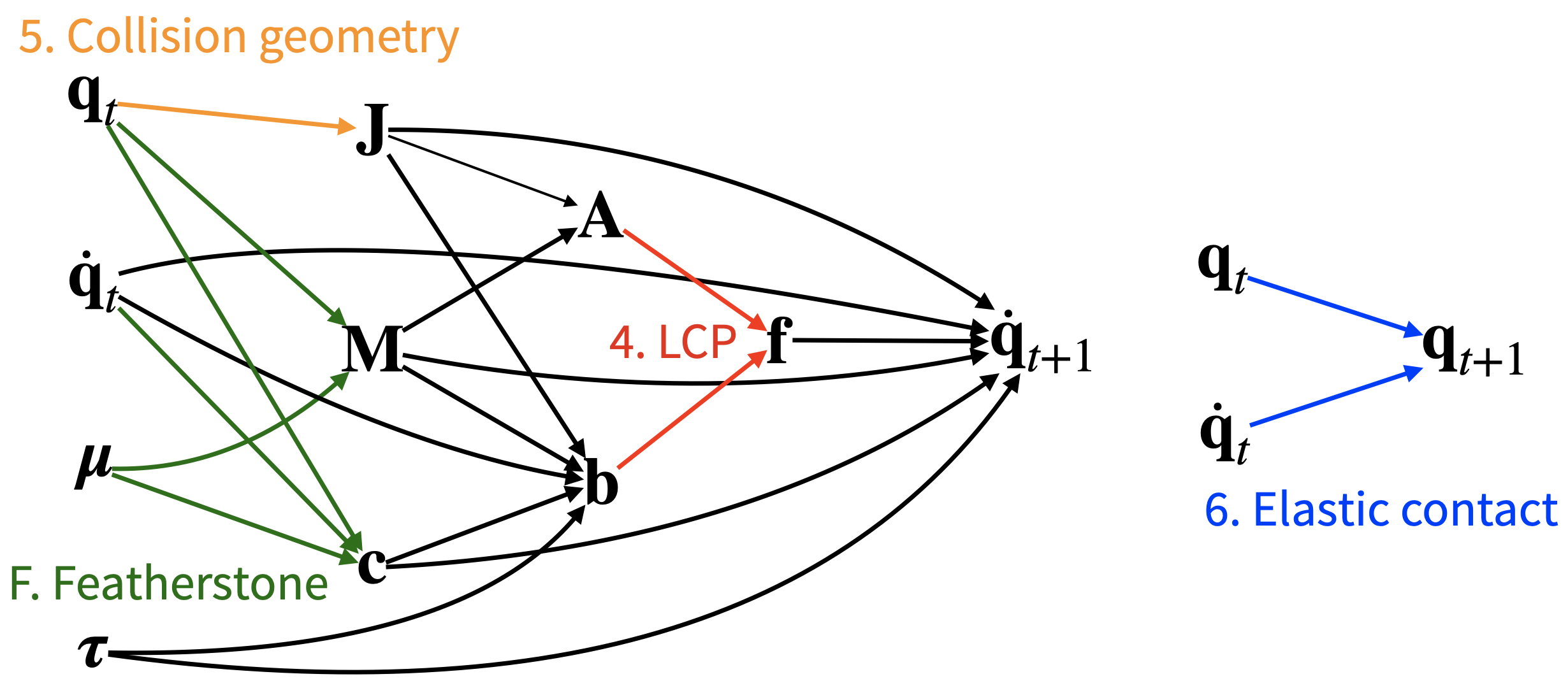}}
\caption{Data flow during a forward simulation, visualizing equations \ref{eqn:EOM} and \ref{eqn:LCP}.
% Each arrow represents a dependency whose gradient computation is described in the corresponding section indicated by the same color.
The inputs are $\vc{q}_t$ and $\dot{\vc{q}}_t$, the current position and velocity in generalized coordinates, $\mu$, the inertial properties, and $\tau$, the external (control) torques on the joints. The outputs are the generalized position and velocity at the next timestep, $\vc{q}_{t+1}$ and $\dot{\vc{q}}_{t+1}$. Every forward arrow represents a dependency in data flow, which must be differentiated during backpropagation. Challenging dependencies, and the relevant sections where we introduce analytical Jacobians, are labeled with colored arrows and text in the diagram.}
\label{fig:data_flow}
\end{center}
\end{figure}

The main task in developing a differentiable physics engine is to solve for the gradient of next velocity $\dot{\vc{q}}_{t+1}$ with respect to the input to the current time step, namely $\vc{q}_t, \dot{\vc{q}}_t, \bm{\tau}_t$, and $\bm{\mu}$. The data flow is shown in Figure \ref{fig:data_flow}. For brevity we refer to the ouput of a function for a given timestep by the same name as the function with a subscript $t$ (e.g. $\bm{J}_t = \bm{J}(\vc{q}_t)$). The velocity at the next step can be simplified to
\begin{equation}
\label{eqn:dynamics_full}
\dot{\bm{q}}_{t+1} = \dot{\bm{q}}_t
+ \bm{M}_t^{-1}\bm{z}_t,
\end{equation}
where $\vc{z}_t \equiv - \Delta t(\bm{c}_t - \bm{\tau}_t) + \bm{J}_t^T\bm{f}_t$. The gradients we need to compute at each time step are written as:
\begin{align}
    \label{eqn:wrt_q}
    \frac{\partial \dot{\bm{q}}_{t+1}}{\partial \bm{q}_t} &=\frac{\partial \bm{M}_t^{-1} \vc{z}_t}{\partial \vc{q}_t} + \bm{M}_t^{-1} \left( - \Delta t\frac{\partial \bm{c}_t}{\partial \vc{q}_t} + \frac{\partial \bm{J}_t^T}{\partial \vc{q}_t}\bm{f}_t \right. \nonumber \\
    & \left. + \vc{J}_t^T\frac{\partial{\vc{f}_t}}{\partial \vc{q}_t} \right) \\ 
    \label{eqn:wrt_dq}
    \frac{\partial \dot{\bm{q}}_{t+1}}{\partial \dot{\bm{q}}_t} &= \bm{I} + \vc{M}_t^{-1} \left(- \Delta t \frac{\partial \vc{c}_t}{\partial \dot{\vc{q}}_t} + \vc{J}_t^T\frac{\partial \vc{f}_t}{\partial \dot{\vc{q}}_t} \right)\\
    \label{eqn:wrt_tau}
    \frac{\partial \dot{\bm{q}}_{t+1}}{\partial \bm{\tau}_t} &= \vc{M}_t^{-1} \left( \Delta t\bm{I} + \vc{J}_t^T\frac{\partial \vc{f}_t}{\partial \vc{\tau}_t} \right)\\
    \label{eqn:wrt_mu}
    \frac{\partial \dot{\bm{q}}_{t+1}}{\partial \bm{\mu}} &= \frac{\partial \vc{M}_t^{-1}\vc{z}_t}{\partial \vc{\mu}} + \vc{M}_t^{-1} \left( -\Delta t\frac{\partial \vc{c}_t}{\partial \vc{\mu}} + \vc{J}_t^T\frac{\partial \vc{f}_t}{\partial \vc{\mu}} \right)
\end{align}

We tackle the tricky intermediate Jacobians in sections that follow. In Section \ref{sec:lcp} we will introduce a novel sparse analytical method to compute the gradients of contact force $\vc{f}_t$ with respect to $\vc{q}_t, \dot{\vc{q}}_t, \bm{\tau}_t, \bm{\mu}$. Section \ref{sec:collision} will discuss $\frac{\partial \vc{J}_t}{\partial \vc{q}_t}$---how collision geometry changes with respect to changes in position. In Section \ref{sec:bounces} we will tackle $\frac{\partial \vc{q}_{t+1}}{\partial \vc{q}_t}$ and $\frac{\partial \vc{q}_{t+1}}{\partial \dot{\vc{q}}_t}$, which is not as simple as it may at first appear, because naively taking gradients through a discrete time physics engine yields problematic results when elastic collisions take place. Additionally, the appendix gives a way to apply the derivations from \cite{kim2012lie} and \cite{carpentier2018analytical} to analytically find $\frac{\partial \vc{M}_t^{-1} \vc{z}_t}{\partial \vc{q}_t}$, $\frac{\partial \vc{c}_t}{\partial \vc{q}_t}$, and $\frac{\partial \vc{c}_t}{\partial \dot{\vc{q}}_t}$.
%TODO: add Jacobians wrt mass back in once they're ready: $\frac{\partial \vc{M}_t^{-1} \vc{z}_t}{\partial \bm{\mu}}$, $\frac{\partial \vc{c}_t}{\partial \bm{\mu}}$

%\js{Q: What does the \textit{sparse} analytical method mean exactly?} \karen{To answer JS's question, we need to read the appendix of de Alvila Belbute-Peres (note to self: having a long last name makes people not want to cite you:). I think the most costly step in their formulation is to solve for Equation 29, which is a large linear system. It's true that we need to solve a linear system too to get the gradient (due to $A_\mathcal{CC}^{-1}$), but it is much more compact submatrix due to the sparsity of the original matrix $A$.}

%The remaining Jacobians which we haven't mentioned are much more pedestrian, and can be found with a trivial application of Matrix Calculus.

%% file: LCP.tex
\section{Differentiating the LCP}
\label{sec:lcp}
This section introduces a method to analytically compute $\frac{\partial \vc{f}}{\partial \vc{A}}$ and $\frac{\partial \vc{f}}{\partial \vc{b}}$. It turns out that it is possible to efficiently get unambiguous gradients through an LCP in the vast majority of practical scenarios, without recasting it as a QP (which throws away sparsity information by replacing the complementarity constraint with an objective function). To see this, let us consider a hypothetical LCP problem parameterized by $\vc{A}, \vc{b}$ with a solution $\vc{f}^{*}$ found during the forward pass: $f_{\text{LCP}}(\bm{A}, \bm{b}) = \bm{f}^{*}$.

For brevity, we only include the discussion on normal contact impulses in this section and leave the extension to friction impulses in Appendix \ref{appendix:friction}. Therefore, each element in $\vc{f}^{*} \geq \vc{0}$ indicates the \emph{normal} impulse of a point contact. By complementarity, we know that if some element $f^{*}_i > 0$, then $v_i = (\vc{A}\vc{f}^{*} + \vc{b})_i = 0$. Intuitively, the relative velocity at contact point $i$ \textit{must be 0} if there is any non-zero impulse being exerted at contact point $i$. We call such contact points ``Clamping" because the impulse $f_i > 0$ is adjusted to keep the relative velocity $v_i = 0$. Let the set $\mathcal{C}$ to be all indices that are clamping. Symmetrically, if $f_j = 0$, then the relative velocity $v_j = (\vc{A}\vc{f}^{*} + \vc{b})_j \geq 0$ is free to vary without the LCP needing to adjust $f_j$ to compensate. We call such contact points ``Separating'' and define the set $\mathcal{S}$ to be all indices that are separating. Let us call indices $j$ where $f_j = 0$ \textit{and} $v_j = 0$ ``Tied.'' Define the set $\mathcal{T}$ to be all indices that are tied.

If no contact points are tied ($\mathcal{T} = \emptyset$), the LCP is strictly differentiable and the gradients can be analytically computed. When some contact points are tied ($\mathcal{T} \neq \emptyset$), the LCP has valid subgradients and it is possible to follow any in an optimization. The tied case is analogous to the non-differentiable points in a QP where an inequality constraint is active while the corresponding dual variable is also zero. In such a case, computing gradients via taking differentials of the KKT conditions will result in a low-rank linear system and thus non-unique gradients \cite{amos2017optnet}.

\subsection{Strictly differentiable cases}
Consider the case where $\mathcal{T} = \emptyset$. We shuffle the indices of $\vc{f}^{*}$, $\vc{v}$, $\vc{A}$ and $\vc{b}$ to group together members of $\mathcal{C}$ and $\mathcal{S}$. The LCP becomes:
\begin{align}
\label{eqn:LCP_formulation}
\begin{matrix*}[l]
\text{find} && \vc{f}^{*}_{\mathcal{C}}, \vc{f}^{*}_{\mathcal{S}}, \vc{v}_{\mathcal{C}}, \vc{v}_{\mathcal{S}}\\
\text{such that} && 
\begin{bmatrix}
\vc{v}_{\mathcal{C}}\\
\vc{v}_{\mathcal{S}}
\end{bmatrix}
=
\begin{bmatrix}
\vc{A}_{\mathcal{CC}} && \vc{A}_{\mathcal{CS}}\\
\vc{A}_{\mathcal{SC}} && \vc{A}_{\mathcal{SS}}\\
\end{bmatrix}
\begin{bmatrix}
\vc{f}^{*}_{\mathcal{C}}\\
\vc{f}^{*}_{\mathcal{S}}
\end{bmatrix}
+
\begin{bmatrix}
\vc{b}_{\mathcal{C}}\\
\vc{b}_{\mathcal{S}}
\end{bmatrix}
\\
&& \vc{f}^{*}_{\mathcal{C}} \geq \vc{0},\; \vc{f}^{*}_{\mathcal{S}} \geq \vc{0}, \; \vc{v}_{\mathcal{C}} \geq \vc{0},\; \vc{v}_{\mathcal{S}} \geq \vc{0} \\
&& \vc{f}_{\mathcal{C}}^{*T} \vc{v}_{\mathcal{C}} = 0,\; \vc{f}_{\mathcal{S}}^{*T} \vc{v}_{\mathcal{S}} = 0.
\end{matrix*}
\end{align}
Since we know the classification of each contact that forms the valid solution $\vc{f}^*$, we rewrite the LCP constraints as follows:
\begin{align}
\begin{matrix*}[l]
\text{find} && \vc{f}^{*}_{\mathcal{C}}, \vc{f}^{*}_{\mathcal{S}}, \vc{v}_{\mathcal{C}}, \vc{v}_{\mathcal{S}}\\
\text{such that} && 
\begin{bmatrix}
\vc{v}_{\mathcal{C}}\\
\vc{v}_{\mathcal{S}}
\end{bmatrix}
=
\begin{bmatrix}
\vc{A}_{\mathcal{CC}} && \vc{A}_{\mathcal{CS}}\\
\vc{A}_{\mathcal{SC}} && \vc{A}_{\mathcal{SS}}\\
\end{bmatrix}
\begin{bmatrix}
\vc{f}^{*}_{\mathcal{C}}\\
\vc{f}^{*}_{\mathcal{S}}
\end{bmatrix}
+
\begin{bmatrix}
\vc{b}_{\mathcal{C}}\\
\vc{b}_{\mathcal{S}}
\end{bmatrix}
\\
&& \vc{f}^{*}_{\mathcal{C}} > \vc{0}, \; \vc{f}^{*}_{\mathcal{S}} = \vc{0}, \; \vc{v}_{\mathcal{C}} = \vc{0},\; \vc{v}_{\mathcal{S}} > \vc{0}.
\end{matrix*}
\end{align}

From here we can see how the valid solution $\vc{f}^{*}$ changes under infinitesimal perturbations $\bm{\epsilon}$ to $\vc{A}$ and $\vc{b}$. Since $\vc{f}^{*}_{\mathcal{S}} = \vc{0}$ and $\vc{v}_{\mathcal{C}} = \vc{0}$, the LCP can be reduced to three conditions on $\vc{f}^{*}_{\mathcal{C}}$:
\begin{align}
\label{eqn:clamping-velocity}
&\vc{0}
=
\vc{A}_{\mathcal{CC}} \vc{f}^{*}_{\mathcal{C}}
+
\vc{b}_{\mathcal{C}} \\
&\vc{f}^{*}_{\mathcal{C}} > \vc{0} \\
\label{eqn:seperation-velocity}
&\vc{A}_{\mathcal{SC}}
\vc{f}^{*}_{\mathcal{C}}
+
\vc{b}_{\mathcal{S}} > \vc{0}.
\end{align}

We will show that these conditions will always be possible to satisfy under small enough perturbations $\bm{\epsilon}$ in the neighborhood of a valid solution. Let us first consider tiny perturbations to $\vc{b}_{\mathcal{S}}$ and $\vc{A}_{\mathcal{SC}}$. If the perturbations are small enough, then Equation \ref{eqn:seperation-velocity} will still be satisfied with our original $\vc{f}_{\mathcal{C}}^{*}$, because we know Equation \ref{eqn:seperation-velocity} already holds \textit{strictly} such that there is some non-zero room to decrease any element of $\vc{A}_{\mathcal{SC}}
\vc{f}^{*}_{\mathcal{C}}
+
\vc{b}_{\mathcal{S}}$ without violating Equation \ref{eqn:seperation-velocity}. Therefore, 
\begin{align}
\frac{\partial \vc{f}_\mathcal{C}^{*}}{\partial \vc{b}_{\mathcal{S}}} = \vc{0} \text{\;\;and\;\;}  \frac{\partial \vc{f}_\mathcal{C}^{*}}{\partial \vc{A}_{\mathcal{SC}}} = \vc{0}.
\end{align}
Next let us consider an infinitesimal perturbation $\bm{\epsilon}$ to $\vc{b}_{\mathcal{C}}$ and the necessary change on the clamping force $\Delta \vc{f}_\mathcal{C}^*$ to satisfy Equation \ref{eqn:clamping-velocity}:
\begin{equation}
\label{eqn:wrt-bc}
\vc{0}
=
\vc{A}_{\mathcal{CC}} (\vc{f}^{*}_{\mathcal{C}} + \Delta \vc{f}^{*}_{\mathcal{C}})
+
\vc{b}_{\mathcal{C}} + \bm{\epsilon}.
\end{equation}

Setting $\vc{A}_{\mathcal{CC}} \vc{f}^{*}_{\mathcal{C}} + \vc{b}_{\mathcal{C}} = \vc{0}$ and assuming $\vc{A}_\mathcal{CC}$ is invertible, the change of the clamping force is given as $\Delta \vc{f}^{*}_{\mathcal{C}} = -\vc{A}_{\mathcal{CC}}^{-1}\bm{\epsilon}$. Since $\vc{f}^{*}_{\mathcal{C}}$ is strictly greater than  $\vc{0}$, it is always possible to choose an $\bm{\epsilon}$ small enough to make $\vc{f}^{*}_{\mathcal{C}} - \vc{A}_{\mathcal{CC}}^{-1}\bm{\epsilon} > 0$ and $\vc{A}_\mathcal{SC} (\vc{f}^*_\mathcal{C} + \Delta \vc{f}^*_\mathcal{C}) + \vc{b}_\mathcal{S} > 0$ remain true. Therefore,
\begin{align}
    \frac{\partial \vc{f}_{\mathcal{C}}^{*}}{\partial \vc{b}_{\mathcal{C}}} = -\vc{A}_{\mathcal{CC}}^{-1}.
\end{align}

Note that $\vc{A}_{\mathcal{CC}}$ is not always invertible because $\vc{A}$ is positive \emph{semidefinite}. We will discuss the case when $\vc{A}_{\mathcal{CC}}$ is not full rank in Section \ref{sec:low-rank-Acc-pseudo} along with a method to stabilize the LCP when there exists multiple LCP solutions in Appendix \ref{appendix:stabilize}.

%\[\frac{\partial f_{\mathcal{S}}^{*}}{\partial b_{\mathcal{C}}} = 0 \]

Lastly, we compute gradients with respect to $\vc{A}_{\mathcal{CC}}$. In practice, changes to $\vc{A}_{\mathcal{CC}}$ only happen because we are differentiating with respect to parameters $\vc{q}$ or $\bm{\mu}$, which also changes $\vc{b}_{\mathcal{C}}$. As such, we introduce a new scalar variable, $x$, which could represent any arbitrary scalar quantity that effects both $\vc{A}$ and $\vc{b}$. Equation \ref{eqn:clamping-velocity} can be rewritten as:
\begin{align}
    \vc{f}^{*}_{\mathcal{C}} = - \vc{A}_{\mathcal{CC}}(x)^{-1}\vc{b}_{\mathcal{C}}(x).
\end{align}

Because $\vc{A}_{\mathcal{CC}}(x)$ and $\vc{b}_{\mathcal{C}}(x)$ are continuous, and the original solution is valid, any sufficiently small perturbation to $x$ will not reduce $\vc{f}^{*}_{\mathcal{C}}$ below 0 or violate Equation \ref{eqn:seperation-velocity}. The Jacobian with respect to $x$ can be expressed as:
\begin{align}
\frac{\partial \vc{f}^{*}_{\mathcal{C}}}{\partial x} = \vc{A}_{\mathcal{CC}}(x)^{-1}\frac{\partial \vc{A}_{\mathcal{CC}}(x)}{\partial x} \vc{A}_{\mathcal{CC}}(x)^{-1}\vc{b}_{\mathcal{C}}(x) \nonumber \\ \label{eqn:force_wrt_x}
+ \vc{A}_{\mathcal{CC}}(x)^{-1}\frac{\partial \vc{b}_{\mathcal{C}}(x)}{\partial x}.
\end{align}
Using $\bm{A} = \bm{J}\bm{M}^{-1}\bm{J}^T$ and $\bm{b} = \bm{J}(\dot{\bm{q}}_t + \Delta t \bm{M}^{-1}(\bm{\tau} - \bm{c}))$, along with the derivations of Featherstone presented in Appendix \ref{appendix:featherstone}, it is possible to compute $\frac{\partial \vc{A}_{\mathcal{CC}}}{\partial x}$ and $\frac{\partial \vc{b}_{\mathcal{C}}}{\partial x}$ for any specific $x$.

\textbf{Remark:} Previous methods \cite{de2018end,liang2019,qiao2020scalable} cast an LCP to a QP and solved for a linear system of size $n+m$ derived from taking differentials of the KKT conditions of the QP, where $n$ is the dimension of the state variable and $m$ is the number of contact constraints. Our method also solves for linear systems to obtain $\bm{A}_{\mathcal{CC}}^{-1}$, but the size of $\bm{A}_{\mathcal{CC}}$ is often much less than $m$ due to the sparsity of the solution $(\bm{f}^*, \bm{v}^*)$.
% \keenon{I changed this from sparsity of $\bm{A}$ to the sparsity of the solution $\bm{f}$, because that connection is much easier to make. Though the sparsity of $\bm{A}$ may lead to sparse solutions $\bm{f}$, it's not obvious (at least to me) that it \textit{must}. The sparsity of $\bm{f}$, however, directly ties to the size of $\bm{A}_{\mathcal{CC}}$. Every element of $\bm{f}$ that is 0 is excluded for $\bm{A}_{\mathcal{CC}}$ by construction.} \karen{Good catch. It's the solution that is sparse, not A.}

\subsection{Subdifferentiable case}

Now let us consider when $\mathcal{T} \neq \emptyset$. Replacing the LCP constraints with linear constraints will no longer work because any perturbation will immediately change the state of the contact and the change also depends on the direction of perturbation. Including the class of ``tied'' contact points to Equation \ref{eqn:LCP_formulation}, we need to satisfy an additional linear system,
\begin{align}
\label{eqn:tie-case}
    \bm{v}_\mathcal{T} &= \bm{A}_\mathcal{TT} \bm{f}_\mathcal{T}^* +  \bm{A}_\mathcal{TC} \bm{f}_\mathcal{C}^* + \bm{A}_\mathcal{TS} \bm{f}_\mathcal{S}^* + \bm{b}_\mathcal{T} \nonumber \\
    &= \bm{A}_\mathcal{TT} \bm{f}_\mathcal{T}^* + \tilde{\bm{b}}_\mathcal{T}, 
\end{align}
where $\tilde{\bm{b}}_\mathcal{T} = \bm{A}_\mathcal{TC} \bm{f}_\mathcal{C}^* + \bm{b}_\mathcal{T}$ and $\bm{v}_\mathcal{T}$ and $\bm{f}_\mathcal{T}^*$ are both zero at the solution. Let $i \in \mathcal{T}$ be the index of a tied contact point. Consider perturbing the $i$'th element of $\tilde{\vc{b}}_\mathcal{T}$ by $\epsilon$. If $\epsilon > 0$, $\bm{A}_\mathcal{TT} \bm{f}_\mathcal{T}^*$ cannot become negative to balance Equation \ref{eqn:tie-case} because $\bm{A}_\mathcal{TT}$ is positive semidefinite and $\bm{f}_\mathcal{T}^*$ must be nonnegative. Therefore, $v_{\mathcal{T}i}$ must become positive, resulting contact point $i$ being separated and $f_{\mathcal{T}i}^{*}$ remaining zero. If $\epsilon < 0$, then $i$ is immediately bumped into the ``clamping'' set $\mathcal{C}$ because $v_{\mathcal{T}i}$ cannot be negative. Therefore, $f^{*}_{\mathcal{T}i}$ must become positive to balance Equation \ref{eqn:tie-case}. The gradients for the clamping and separating cases are both valid subgraidents for a tied contact point. In an optimization, we can choose either of the two subgradients at random without impacting the convergence \cite{boyd2003subgradient}. In practice, encountering elements in $\mathcal{T}$ is quite rare for practical numerical reasons.

%The conditions for this contact point can be written as
%\[A_{ii}f^*_i + \sum_{j\neq i}A_{ij}f^*_j + b_i = v_i, \text{\;where\;} f_i^* = 0,\;\; v_i = 0
%\]
%Consider perturbing $b_i$ by $\epsilon$. If $\epsilon > 0$, $v_i$ must be   

%You can think of elements in $\mathcal{T}$ as being just like ReLU units that are at exactly 0. The derivative is technically undefined, but it's perfectly fine in a practical optimization to tie break to either of the two subgradients available to you at random, and you're still guaranteed to converge \cite{boyd2003subgradient}. In any case, encountering elements in $\mathcal{T}$ is about as rare as encountering ReLU's at exactly 0 in the wild for practical numerical reasons.

%With that, we can now efficiently compute $\frac{\partial f}{\partial A}$ and $\frac{\partial f}{\partial b}$.

\subsection{When \texorpdfstring{$\vc{A}_{\mathcal{CC}}$}{A\_cc} is not full rank}
\label{sec:low-rank-Acc-pseudo}
When $\vc{A}_{\mathcal{CC}}$ is not full rank, the solution to $f_{\text{LCP}}(\vc{A}, \vc{b}) = \vc{f}^{*}$ is no longer unique. Nevertheless, once a solution is computed using any algorithm and the clamping set $\mathcal{C}$ is found, we can use the stabilization method proposed in Appendix \ref{appendix:stabilize} to find the least-squares minimal $\vc{f}^{*}$ that solves the LCP. The gradient of clamping forces can then be written as:
\begin{align}
\frac{\partial \vc{f}^{*}_{\mathcal{C}}}{\partial x} = \vc{A}_{\mathcal{CC}}^{-1}\frac{\partial \vc{A}_{\mathcal{CC}}}{\partial x} \vc{A}_{\mathcal{CC}}^{-1}\vc{b}_{\mathcal{C}} \nonumber + \vc{A}_{\mathcal{CC}}^{-1}\frac{\partial \vc{b}_{\mathcal{C}}}{\partial x}\\ \label{eqn:contact_wrt_Acc}
+ (\vc{I} - \vc{A}^+_\mathcal{CC}\vc{A}_\mathcal{CC})\big(\frac{\partial \vc{A}_\mathcal{CC}^T}{\partial x}\big) \vc{A}_\mathcal{CC}^{+T}\vc{A}_\mathcal{CC}^+ ,
\end{align}
where $\vc{A}_\mathcal{CC}^+$ is the pseudo inverse matrix of the low-rank $\vc{A}$. For numerical stability, we can solve a series of linear systems instead of explicitly evaluating $\vc{A}_\mathcal{CC}(x)^+$.
% \begin{align*}
%     \vc{A}_{\mathcal{CC}}(x)\vc{y}_1 &= \vc{b}_{\mathcal{C}}(x), \\
%     \vc{A}_{\mathcal{CC}}(x)\vc{y}_2 &= \frac{\partial \vc{b}_{\mathcal{C}}(x)}{\partial x}, \\
%     \vc{A}_{\mathcal{CC}}(x)\vc{y}_3 &= \frac{\partial \vc{A}_{\mathcal{CC}}(x)}{\partial x} \vc{y}_1.
% \end{align*}

% Once we solve for $\vc{y}_1$, $\vc{y}_2$ and $\vc{y}_3$, we can compute $\frac{\partial \vc{f}^{*}_{\mathcal{C}}}{\partial x}$ by solving the final linear system:
% \[
% \vc{A}_{\mathcal{CC}}(x) \frac{\partial \vc{f}^{*}_{\mathcal{C}}}{\partial x} = \vc{y}_3 + \vc{y}_2.
% \]

\subsection{Complementarity-aware gradients via contact constraints}

\label{sec:complementarity-aware}

Sometimes analytically correct Jacobians through the LCP can actually prevent an optimizer from finding a good solution. When a contact $i$ is clamping ($i \in \mathcal{C}$), we effectively impose a constraint that the relative velocity at that contact point is zero no matter how we push \textit{or pull} on the contact. This prevents the gradients from pointing towards any motions that require breaking contact, because our constraints will zero out gradients that lead to $\vc{v}_i > 0$. %Clamped contacts may as well be new joints, from the perspective of the gradient. %

This behavior is caused by the complementarity constraint requiring that at most one of $v_i$ or $f_i$ can be non-zero. This phenomenon grows out of the short-sightedness of the gradient (it only considers an infinitely small neighborhood $\epsilon$ around the current state). While it is true that small attempts to push $v_i > 0$ will result in no change to $\vc{v}$ as $\vc{f}$ compensates to keep $v_i = 0$, eventually we will reach a discontinuity where $\vc{f}$ can no longer prevent the contact from separating. However, a gradient-based optimizer may never take steps in that direction because gradients in that direction are 0. In this section we propose a heuristic to opportunistically explore ``projecting forward'' to the discontinuity where the complementarity constraint flips during backpropagation, depending on the gradient of loss function with respect to $\vc{v}$ and $\frac{\partial \ell}{\partial \vc{v}}$. 

% Karen's original intro paragraphs:

% Analytically correct gradients through the LCP are critical, but they do not necessarily improve the convergence of downstream optimization problems because the gradient information sometimes could be myopic due to discontinuous contact states. However, by analyzing contact dynamics, we can compute heuristic-based ``complementarity-aware gradients,'' that (while no longer valid gradients) can help the optimizer escape saddle points.

% When taking gradients through a contact point, there are two scenarios in which the gradients could be myopic. The first case arises when the agent/system attempts to change the contact velocity using its own control force for a contact point initialized as "clamping". The second case arises when the agent attempts to change the contact force for a contact point initialized as "separating". In both scenarios, the gradients are sensitive to initialization because they are limited by the LCP constraints and their initial contact state classification.

\begin{figure}[ht]
\begin{center}
\centerline{\includegraphics[width=\columnwidth]{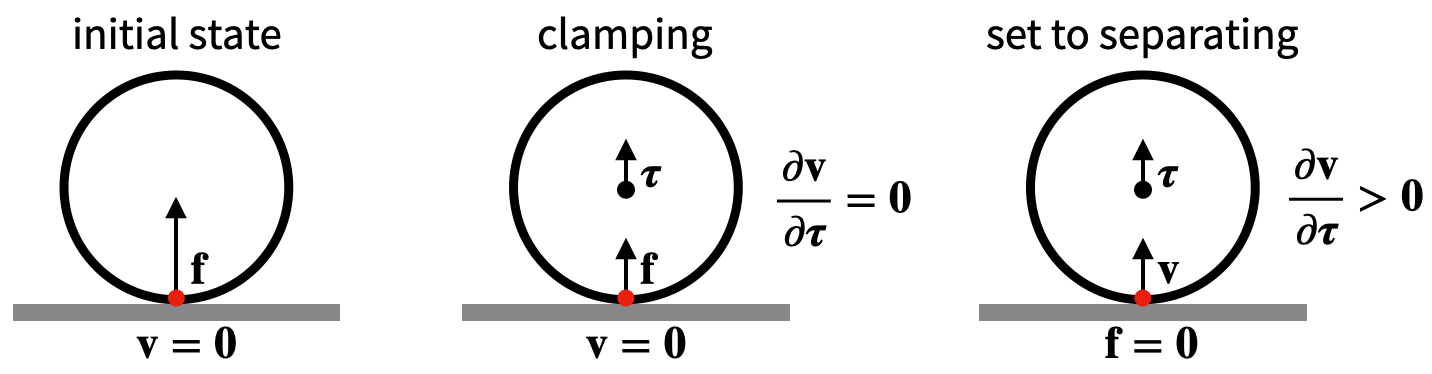}}
\caption{Increasing force from the linear actuator ($\bm{\tau}$) is met by an equal and opposite decrease in ground reaction force $\vc{f}$, resulting in no change in velocity and thus no gain in height. Gradients of loss function will be zeroed out by $\frac{\partial \vc{v}}{\partial \bm{\tau}} = \vc{0}$ when backpropgating through the contact. In contrast, if the contact is classified as ``separating", the gradient will increase the linear actuator, resulting in an upward force $\vc{v}$.}
\label{fig:actuation_strategy}
\end{center}
\end{figure}

To make it more concrete, let us consider a simple example of a 2D circle attached to a linear actuator that can produce force along the vertical axis (Figure \ref{fig:actuation_strategy}). Our goal is to lift the circle up to a target height above the ground by generating an upward velocity using the linear actuator. When the circle is resting on the ground under gravity, the contact point with the ground is classified as ``clamping''. This means that any tiny perturbation in the control force of linear actuator will be met with an exactly offsetting contact force to ensure that the relative velocity between the circle and the ground remains zero. As such, no matter what gradient of loss function with respect to the control force we try to backpropagate through the contact, we will always get $\frac{\partial l}{\partial \bm{\tau}} = \vc{0}$. 

\begin{figure}[ht]
\begin{center}
\centerline{\includegraphics[width=\columnwidth]{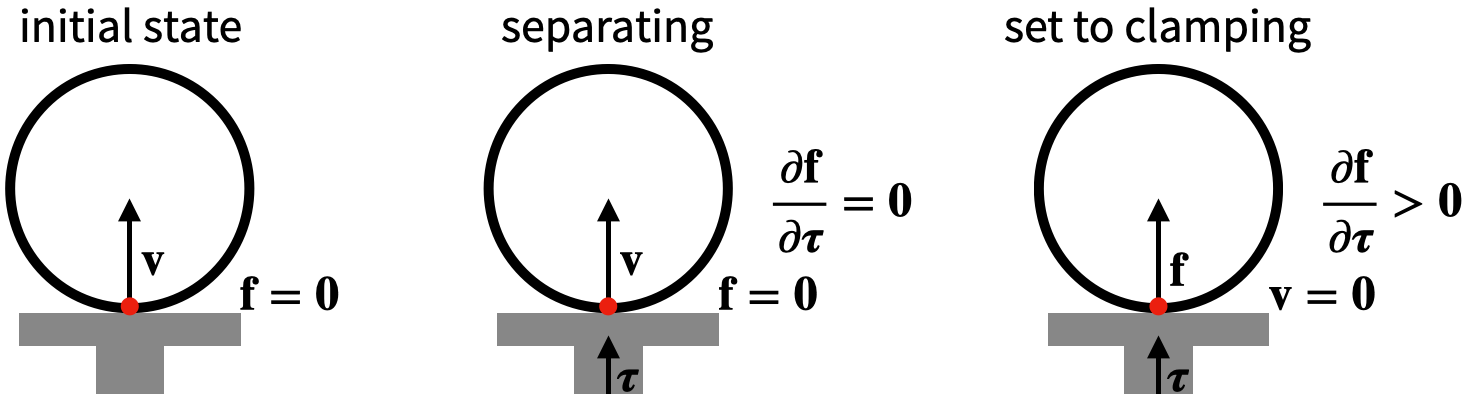}}
\caption{If the initial contact state is ``separating'', the platform cannot apply contact force to the circle, resulting in zero gradient of the contact force with respect to the control force of the platform, $\frac{\partial \bm{f}}{\partial \bm{\tau}} = \vc{0}$. However, if the contact is classified as ``clamping'', we can utilize the contact force from the platform to push the circle upwards by increasing the control force of the platform. }
\label{fig:normal_force_strategy}
\end{center}
\end{figure}

Consider another example where the same 2D circle is uncontrolled and resting on a linearly actuated platform (Figure \ref{fig:normal_force_strategy}). To lift the circle up, we need to utilize the contact force from the platform. If the initialization of the problem results in the contact point being classified as ``separating", no constraint force will be applied on the contact point, as if the contact did not exist. Therefore, the gradient of contact force with respect to the platform control force is zero, which may cause the optimization to be trapped in a saddle point.

In both examples, the issues can be resolved by classifying the contact point differently. If we know that to solve this problem we need to actuate the circle and the platform separately, we would set the contact point to be ``separating.'' This gives the optimizer a chance to explore a solution with contact breaking. On the other hand, if we know that we need to exploit the contact force between the circle and the platform, we would relabel a separating contact point to be ``clamping''. This lets the optimizer search for the most effective contact force that reduces the loss function.

Is there a way to know which strategy to use in advance? In the general case, the answer seems to be no, but we can do better than naive gradients which stick with whatever contact classification they were initialized into and do not explore to break out of resulting saddle points. We propose a heuristic to explore more broadly at a constant additional cost.

We propose that always picking the contact strategy that results in the largest $||\frac{\partial \ell}{\partial \dot{\vc{q}}_t}||_2^2 + \frac{||\frac{\partial \ell}{\partial \bm{\tau}_t}||_2^2}{\Delta t}$ is a good heuristic during learning for avoiding saddle points. While we could try backprop through all $\mathcal{O}(2^n)$ possible strategies and pick the best one, that gets expensive as the number of contacts becomes larger than a small handful. Instead of exhaustively searching for all possible combinations of contact states with exponential complexity, we propose to only check two classifications: 

% When the gradient of the loss function becomes zero,  $\frac{\partial \ell}{\partial \dot{\vc{q}}_t} = \vc{0}$ and $\frac{\partial \ell}{\partial \bm{\tau}_t} = \vc{0}$, it is possible that we hit a saddle point for the current strategy and there exists another contact classification that could reduce the loss function further. That is, we could find a larger $||\frac{\partial \ell}{\partial \dot{q}_t}||_2^2 + ||\frac{\partial \ell}{\partial \tau_t}||_2^2$. Instead of exhaustively searching for all possible combinations of contact states with an exponential complexity, we propose only check two classifications: 
\begin{enumerate}
    \item First, check the ``correct'' contact classification solved by the LCP.
    \item Second, we compute $\frac{\partial \ell}{\partial \vc{v}}$ (the gradient of loss with respect to relative contact velocity), and use the elements of $\frac{\partial \ell}{\partial \vc{v}}$ to compute the ``clamping'' and ``separating'' sets as follows. Take any indices $i$ that are trying to increase the relative velocity at that contact $\frac{\partial \ell}{\partial v_i} < 0$, which implies the optimizer is feebly trying to separate the objects, and put those indices into ``separating'' to remove the constraint that $v_i = 0$. Take any indices $i$ where $\frac{\partial \ell}{\partial v_i} > 0$, which implies the optimizer is trying to move the objects closer together (which would violate contact constraints), and put them into ``clamping'' to impose the constraint that $v_i = 0$.
\end{enumerate}

The contact strategy with the larger $||\frac{\partial \ell}{\partial \dot{\vc{q}}_t}||_2^2 + \frac{||\frac{\partial \ell}{\partial \bm{\tau}_t}||_2^2}{\Delta t}$ is used during backpropagation for learning. We call the gradient produced by this exploratory procedure a ``Complementarity-aware Gradient,'' and find empirically that it can help avoid saddle points during trajectory optimization. We show an example of this in Section \ref{sec:results}. 

The complementarity-aware gradients do not guarantee to improve global convergence because the gradient is chosen for each contact point independently, which might not result in an aggregated gradient $\frac{\partial \ell}{\partial \bm{\tau}}$ that moves in a globally optimal direction. However, if an optimization problem currently gets stuck in a saddle point, complementarity-aware gradients provide another tool for practitioners to try.

%% file: collision_geometry.tex
\section{Gradients through collision geometry}
\label{sec:collision}
%\karen{Can all this be done by auto diff?} \keenon{Yes, in principle. It would require an enormous and wasteful computation, but there's nothing not trivially differentiable (like an optimization problem) preventing it. The benefit of doing it analytically is getting to shortcut a lot of computation by putting everything in world space, rather than passing every gradients through a long chain of predictable joint transforms.}
This section addresses efficient computation of $\frac{\partial \vc{J}_t^T\vc{f}}{\partial \vc{q}_t}$, the relationship between position and joint impulse. In theory, we could utilize auto-diff libraries for the derivative computation. However, using auto-diff and passing every gradient through a long kinematic chain of transformations is inefficient for complex articulated rigid body systems. In contrast, computing the gradients symbolically shortcuts much computation by operating directly in the world coordinate frame.

% Because $\frac{\partial J_t}{\partial q_t}$ is a 3-dimensional tensor, and we only ever use it as $\frac{\partial J_t}{\partial q_t}^Tf$ for some arbitrary vector $f$, we'll actually compute a 2-dimensional Jacobian instead in order to keep the math manageable:

% \begin{equation}
%     \frac{\partial J_t^Tf}{\partial q_t}
% \end{equation}

% The quantity $J_t^Tf$ gives us joint impulses resulting from applying a fixed vector of impulse magnitudes $f$ at all the contact points along the contact normals. Without loss of generality, let's consider how the $i$'th DOF joint impulses are generated.

Let $\mathcal{A}_i \in se(3)$ be the screw axis for the $i$'th DOF, expressed in the world frame. Let the $k$'th contact point give an impulse $\mathcal{F}_k \in dse(3)$, also expressed in the world frame. Let $\psi_{i,k} \in \{-1, 0, 1\}$ be the relationship between contact $k$ and joint $i$. $\psi_{i,k} = 1$ or $\psi_{i,k} = -1$ means contact $k$ is on a child body of joint $i$. Otherwise, $\psi_{i,k} = 0$. The total joint impulse caused by contact impulses for the $i$'th joint is given by:

\begin{equation}
\label{eqn:joint-impulse}
(\vc{J}_t^T\vc{f})_i = \sum_k \psi_{i,k} \mathcal{A}_i^T \mathcal{F}_k = \mathcal{A}_i^T \sum_k \psi_{i,k} \mathcal{F}_k.
\end{equation}

Taking the derivative of Equation \ref{eqn:joint-impulse} gives
\begin{equation}
\frac{\partial (\vc{J}_t^T\vc{f})_i}{\partial \vc{q}_t} = \frac{\partial \mathcal{A}_i}{\partial \vc{q}_t}^T \sum_k \psi_{i,k}\mathcal{F}_k + \mathcal{A}_i^T \sum_k \psi_{i,k}\frac{\partial \mathcal{F}_k}{\partial \vc{q}_t}.
\end{equation}

Evaluating $\frac{\partial \mathcal{A}_i}{\partial \vc{q}_t}$ is straightforward, but computing $\frac{\partial \mathcal{F}_k}{\partial q_t}$ requires understanding how the contact normal $\vc{n}_k \in \mathcal{R}^3$ and contact position $\vc{p}_k \in \mathcal{R}^3$ change with changes in $\vc{q}_t$. 

Let $\mathcal{F}_k$ be a concatenation of torque and force in $dse(3)$. The derivative with respect to the current joint position $\vc{q}_t$ is:

% \begin{equation}
% \mathcal{F}_k = \begin{bmatrix} 
%  \vc{n}_k \times \vc{p}_k \\
%  \vc{n}_k
% \end{bmatrix}
% \end{equation}

% That means:

\begin{equation}
\frac{\partial \mathcal{F}_k}{\partial \vc{q}_t} = \begin{bmatrix} \frac{\partial \vc{n}_k}{\partial \vc{q}_t} \times \vc{p}_k + \vc{n}_k \times \frac{\partial \vc{p}_k}{\partial \vc{q}_t}\\
\frac{\partial \vc{n}_k}{\partial \vc{q}_t}
\end{bmatrix}
\end{equation}

% Once this is understood (i.e. once we know $\frac{\partial \vc{n}_k}{\partial q_t}$ and $\frac{\partial \vc{p}_k}{\partial q_t}$), the math to compute $\frac{\partial \mathcal{F}_k}{\partial q_t}$ is predictable but tedious, and we refer the reader to our published code for a complete reference.

% DiffDart supports collisions between spheres, capsules, boxes, and arbitrary convex meshes. The reason it's important to support primitive shapes such as spheres and capsules explicitly, in addition to convex primitives, is that primitive shapes with curved surfaces like spheres and capsules have different gradients through their contact normals $\frac{\partial \vc{n}_k}{\partial q_t}$ than a mesh-based approximation.

% To allow gradients to exist, our collision detection routines must return \textit{exact} contact points and normals. This means algorithms that return only approximate contact position and normal, like the Gilbert-Johnson-Keerthi (GJK) algorithm or Minkowski Portal Refinement (MPR) algorithm, need to have their outputs ``stabilized'' through a post-processing step with exact semantics. For example, in mesh-mesh collisions caused by a vertex intersecting a face, we enforce that the collision point always be exactly at the vertex, and the collision normal always be exactly the normal of the face.

\begin{figure}[ht]
\begin{center}
\centerline{\includegraphics[width=\columnwidth]{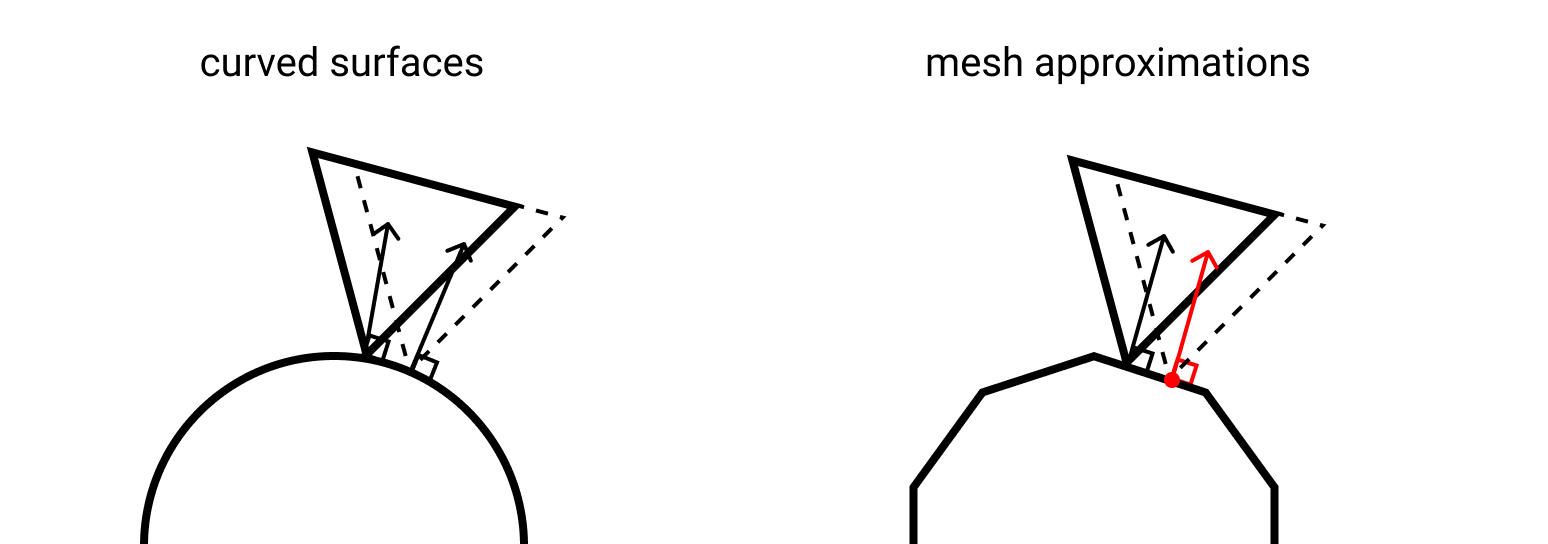}}
\caption{$\frac{\partial \vc{n}_k}{\partial \vc{q}}$ can be set to zero for mesh or polygonal geometry, but has to be computed analytically for curved geometry.}
\label{fig:curved-vs-mesh}
\end{center}
\end{figure}

It turns out that computing $\frac{\partial \vc{n}_k}{\partial q_t}$ for curved primitives shape colliders (spheres and capsules) requires different treatment from meshes or polygonal primitives. To understand why, consider using a high polycount mesh to approximate a true sphere as shown in Figure \ref{fig:curved-vs-mesh}.

On a mesh approximation of a sphere, infinitesimally moving the contact point $\vc{p}_k$ by $\epsilon$ (by perturbing $\vc{q}$ of the triangle) will not cause the normal of the contact face $\vc{n}_k$ to change at all, because we remain on the same face of the mesh. So $\frac{\partial \vc{n}_k}{\partial \vc{q}}$ is always zero for a mesh or polygonal shape. However, on a true sphere, an infinitesimal perturbation of the contact point with the sphere (no matter how small) will cause the contact normal with the sphere $\vc{n}_k$ to change. The way the contact normal $\vc{n}_k$ changes with position (e.g. $\frac{\partial \vc{n}_k}{\partial \vc{q}}$) is often crucial information for the optimizer to have in order to solve complex problems, and mesh approximations to curved surfaces falsely set $\frac{\partial \vc{n}_k}{\partial \vc{q}} = 0$.

%Mesh approximations to curved surfaces produce incorrect optimizations, because we get the wrong Jacobian of contact normal $\frac{\partial \vc{n}_k}{\partial q_t}$.
% \karen{Is the main point that $\frac{\partial \vc{n}_k}{\partial q_t}$ is hard to compute for curved primitive shapes?}

We refer the interested reader to Appendix \ref{appendix:contact} for a discussion of how we compute $\frac{\partial \vc{n}_k}{\partial \vc{q}_t}$ and $\frac{\partial \vc{p}_k}{\partial \vc{q}_t}$ for different combinations of collider types.

%% file: elastic_contact.tex
\section{Gradients through elastic contacts}
\label{sec:bounces}

\begin{figure}[ht]
\begin{center}
\centerline{\includegraphics[width=\columnwidth]{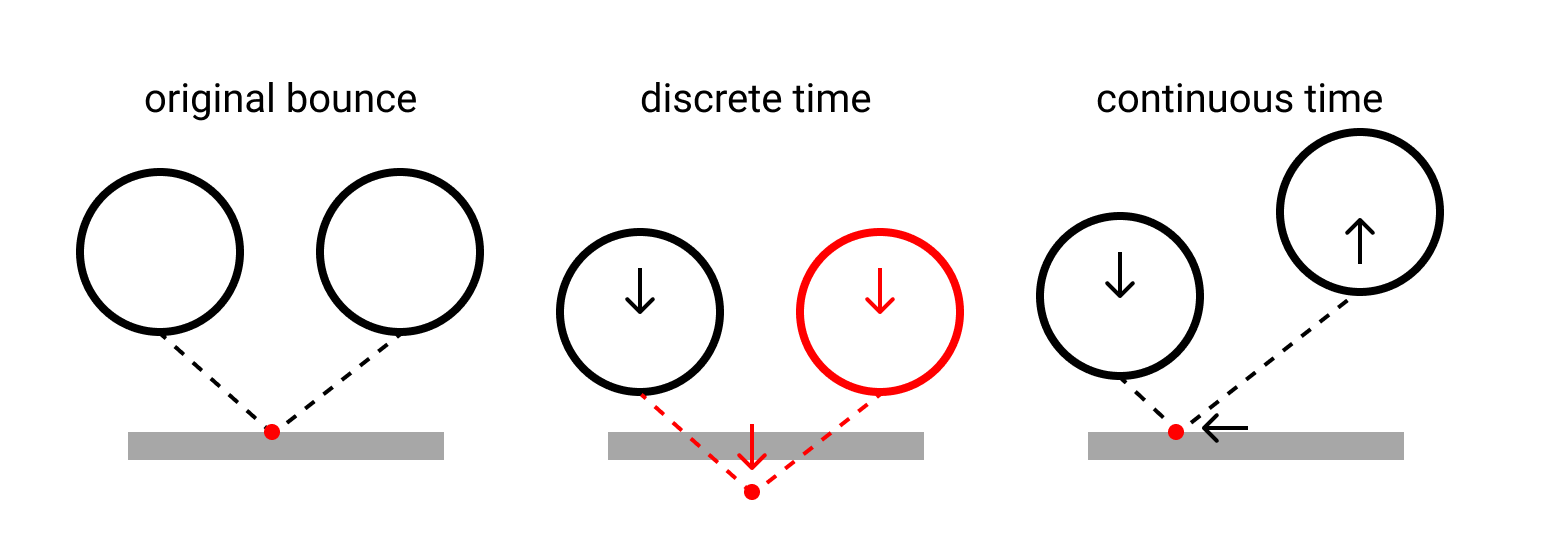}}
\caption{Discrete-time Jacobians do not adequately describe the dynamics of an elastic collision. In the system pictured here, letting $\vc{q}$ be a scalar giving the distance from the ground, discrete-time would lead us to falsely believe that $\frac{\partial \vc{q}_{t+1}}{\partial \vc{q}} = 1$. In continuous time, we would instead expect $\frac{\partial \vc{q}_{t+1}}{\partial \vc{q}} = -\sigma$, where $\sigma$ is the coefficient of restitution. Modeling this requires understanding how the time of collision changes with initial conditions, and how that affects the final state after the step.}
\label{fig:bounce}
\end{center}
\end{figure}

DiffTaichi \cite{difftaichi} pointed out an interesting problem that arises from discretization of time in simulating elastic bouncing phenomenon between two objects. The problems arise from the discrete time integration of position: $\vc{q}_{t+1} = \vc{q}_t + \Delta t \dot{\vc{q}}_t$. The Jacobians $\frac{\partial \vc{q}_{t+1}}{\partial \vc{q}_t} = \vc{I}$ and $\frac{\partial \vc{q}_{t+1}}{\partial \dot{\vc{q}}_t} = \Delta t \vc{I}$ are correct for most scenarios. However, when an elastic collision occurs, the discrete time integration scheme creates problems for differentiation. In a discrete time world, the closer the object is to the collision site at the beginning of the time step when collision happens, the closer to the collision site it ends up at the end of that time step. In continuous time, however, the closer the object begins to its collision site, the \textit{further away it ends up}, because the object changes velocity sooner (Figure \ref{fig:bounce}).

DiffTaichi \cite{difftaichi} proposed using continuous-time collision detection and resolution during training, but noted that switching to discrete-time at test time did not harm performance. Since introducing full continuous-time collision detection is computationally costly for complex dynamic systems and scenes, we instead opt to find a $\frac{\partial \vc{q}_{t+1}}{\partial \vc{q}_t}$ and $\frac{\partial \vc{q}_{t+1}}{\partial \dot{\vc{q}}_t}$ that approximates the behavior of continuous-time Jacobians when elastic collisions occur.

Let $\vc{p}_{t,i}$ be the relative distance at contact $i$ at time $t$ and $\vc{v}_{t,i}$ be the relative velocity. We further define $\sigma_i$ as the coefficient of restitution at contact $i$ and $t + \vc{c}_i$ as the time of collision of contact $i$, where $\vc{c}_i \in \mathcal{R}$. Assuming the relative velocity is constant in this time step, we get $\vc{c}_i = -\vc{p}_i / \vc{v}_i$ and $\vc{p}_{t+1,i} = (\Delta t - \vc{c}_i)\sigma \vc{v}_{t,i}$. From there we have:

%Our approach is to set up continuous time equations for each collision independently, find their gradients, and then attempt to find a pair of Jacobians $\frac{\partial \vc{q}_{t+1}}{\partial \vc{q}_t}$ and $\frac{\partial \vc{q}_{t+1}}{\partial \dot{\vc{q}}_t}$ that approximate our desired behavior at each of the contact points as closely as possible.

%First, we find a continuous time version of $\frac{\partial \vc{p}_{t+1,i}}{\partial \vc{p}_{t,i}}$ and $\frac{\partial \vc{p}_{t+1,i}}{\partial \vc{v}_{t,i}}$. To do this, first define $\sigma_i$ as the coefficient of restitution at contact $i$, and $t + \vc{c}_i$ as the time of collision of contact $i$, where $\vc{c}_i \in \mathcal{R}$. We know $\vc{c}_i = -\vc{p}_i / \vc{v}_i$. Then we also know that $\vc{p}_{t+1,i} = (\Delta t - \vc{c}_i)\sigma \vc{v}_{t,i}$. From there, we have:

\begin{equation}
\frac{\partial \vc{p}_{i,t+1}}{\partial \vc{p}_{i,t}} = -\sigma_i, \;\;\;\frac{\partial \vc{p}_{i,t+1}}{\partial \vc{v}_{i,t}} = -\sigma_i\Delta t.
\end{equation}

After finding the above gradients for each collision independently, we need to find a pair of Jacobians, $\frac{\partial \vc{q}_{t+1}}{\partial \vc{q}_t}$ and $\frac{\partial \vc{q}_{t+1}}{\partial \dot{\vc{q}}_t}$, that approximate our desired behavior at each of the contact points as closely as possible.

\begin{align}
\label{eqn:jacobian_for_bounces}
\frac{\partial \vc{p}_{i,t+1}}{\partial \vc{p}_{i,t}} &=
\frac{\partial \vc{p}_{i,t+1}}{\partial \vc{q}_{t+1}} \frac{\partial \vc{q}_{t+1}}{\partial \vc{q}_{t}}\frac{\partial \vc{q}_{t}}{\partial \vc{p}_{i,t}} \nonumber \\
&= \vc{J}_{i,t+1} \frac{\partial \vc{q}_{t+1}}{\partial \vc{q}_t} \vc{J}_{i,t}^{-1}, \;\mathrm{for}\;i = 1\cdots m
\end{align}
where $\vc{J}_{i,t+1}$ is the Jacobian matrix of contact $i$, $\vc{J}_{i,t}^{-1}$ is the pseudo inverse Jacobian, and $m$ is the total number of contact points. Our goal is to find a Jacobian $\frac{\partial \vc{q}_{t+1}}{\partial \vc{q}_t}$ that satisfies the above equations (Equation \ref{eqn:jacobian_for_bounces}) as closely as possible (details in Appendix \ref{appendix:bounces}). Once we find a satisfactory approximation for $\frac{\partial \vc{q}_{t+1}}{\partial \vc{q}_t}$, we can get
\begin{equation}
\frac{\partial \vc{q}_{t+1}}{\partial \vc{\dot{q}}_t} = \Delta t \frac{\partial \vc{q}_{t+1}}{\partial \vc{q}_t}.
\end{equation}

%% file: evaluation.tex
\begin{table}[t]
\caption{Benchmarks against Finite Differencing}
\label{engine-comparison-short}
\vskip 0.1in
\begin{center}
\begin{small}
\begin{sc}
\begin{tabular}{l c c c}
\toprule
Environment & Analytical & Central  & Speedup \\
 & & Differences & \\
\midrule
Atlas  & 16.1ms & 737ms & 45.8x  \\
Half Cheetah  & 0.870ms & 7.52ms & 8.64x  \\
Jump-Worm & 0.484ms & 3.08ms & 6.36x \\
Catapult & 0.576ms & 3.69ms & 6.40x \\
\bottomrule
\end{tabular}
\end{sc}
\end{small}
\end{center}
\vskip -0.1in
\end{table}

\section{Evaluation}
\label{sec:results}

First, we evaluate our methods by testing the performance of gradient computation. In addition, we compare gradient-based trajectory optimization enabled by our method to gradient-free stochastic optimization, and discuss implications. We also demonstrate the effectiveness of the complementarity-aware gradients in a trajectory optimization problem. Finally, we show that our physics engine can solve optimal control problems for complex dynamic systems with contact and collision, including an Atlas humanoid jumping in the air.

\subsection{Performance}
Controlled comparison to existing methods can be challenging because they use different formulations for forward simulation, essentially simulating different physical phenomena. We therefore benchmark the performance of our method at the atomic level--measuring the computation time of a Jacobian for a single time step using a single-core CPU and comparing it to finite differencing methods, using the same forward simulation process provided by an existing non-differentiable physics engine (DART). This comparison removes factors due to differences in forward simulation, implementation techniques, and computation resources, and focuses on the speed gain solely contributed by our gradient computation method.

Table \ref{engine-comparison-short} contains abbreviated benchmark performance of our Jacobians against central differencing. We evaluate our results in four environments: the Atlas robot on the ground (33 DOFs, 12 contact points), Half Cheetah on the ground (9 DOFs, 2 contact points), Jump-Worm (5 DOFs, 2 contact points), and Catapult (5 DOFs, 2 contact points). For each environment, we compare the speed of evaluating all five primary Jacobians. For a complete table, including the speed of individual Jacobian evaluations, see Appendix \ref{appendix:benchmarks}. % Our method clearly outperforms finite differencing in both speed and accuracy across a range of environments. \karen{Should we add a column to show Ridder FD?} \keenon{Since we're already short on space to explain stuff in the main paper, I think it's probably easier to leave it to the appendix. Though we should revisit this after the bugs are fixed and we have JS's latest optimizations fully integrated, which will get us at least 88x faster than finite differencing}

%The most complex environment, Atlas, demonstrates the greatest speedup, with an overall 45x speedup. Even in simple environments like Jump-Worm, our method gives a significant increase in speed and accuracy.

\subsection{Gradient-based vs gradient-free trajectory optimization}

\begin{figure}[ht]
\label{fig:ddp}
\begin{center}
\centerline{\includegraphics[width=\columnwidth]{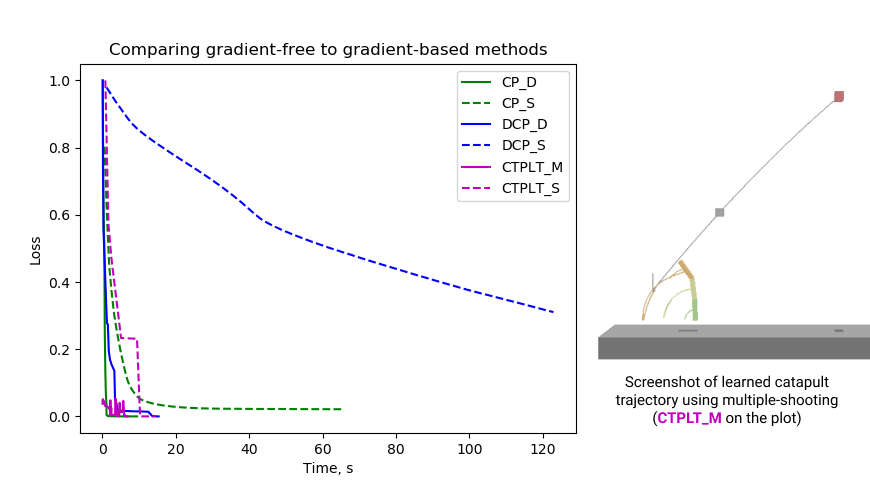}}
\caption{Comparing wall clock time to find a single-pendulum cartpole (CP), double-pendulum cartpole (DCP), and catapult (CTPLT) trajectory, using DDP (D), SS (S), and Multiple Shooting (M). The results are unsurprising: gradient information speeds convergence tremendously.}
\label{fig:gradients}
\end{center}
\end{figure}

To demonstrate the benefit of analytical Jacobians of dynamics, we compare trajectory optimization on our catapult trajectory problem between Multiple Shooting and Stochastic Search (SS) \cite{boutselis2020constrained}, as well as cartpole and the double-pendulum using both Differential Dynamic Programming (DDP) \cite{tassa2012synthesis,jacobson1970differential,todorov2005generalized} and Stochastic Search \cite{boutselis2020constrained}. The gradient-based methods (DDP and Multiple Shooting) are able to converge much more quickly, because the additional convergence gained from analytical Jacobians more than offsets the time to compute them. See Figure \ref{fig:ddp}. 

% \karen{Can we also add Jumpworm comparison? Can we take a log on the height reward so that SS does not get infinite reward from sending the worm to Mars? I think there’s great value to include this example because it involves contact. Even at the end, both DDP and SS get good results and DDP does not beat SS, it is already breaking new ground by solving a contact-rich problem using DDP.}

\subsection{Complementarity-aware gradients}
To highlight the complementarity-aware gradients, we solve a trajectory optimization problem of a drone taking-off from the ground and reaching a fixed height in $500$ timesteps.

Because the drone is initialized resting on the ground, it has a contact with the ground that is classified as ``clamping.'' When we attempt to optimize the control on the drone using correct gradients, we get zero gradients and make no progress. By contrast, when we use our \textit{complementarity aware gradient}, while we still see no change in loss for the first few iterations of SGD, we're able to get non-zero gradients and escape the saddle point (Figure \ref{fig:gradients}). See the supplementary video for the resulting drone trajectories.

%This means that the optimizer sees the drone and the ground as basically attached with a joint, and the optimizer is only able to explore trajectories that attempt to exploit the normal force between the drone and the ground. Because it's impossible to achieve the goal using that strategy, this means we get stuck in a saddle point.

%When we attempt to optimize the control on the drone using correct gradients, we get zero gradients and make no progress. See the blue line in Figure \ref{fig:gradients}. By contrast, when we use our \textit{complementarity aware gradient}, while we still see 0 change in loss for the first few iterations of SGD, we're able to get non-zero gradients and escape the saddle point. See the orange line in Figure \ref{fig:gradients}. See the supplementary video for the resulting drone trajectories.

\begin{figure}[ht]
\begin{center}
\centerline{\includegraphics[width=\columnwidth]{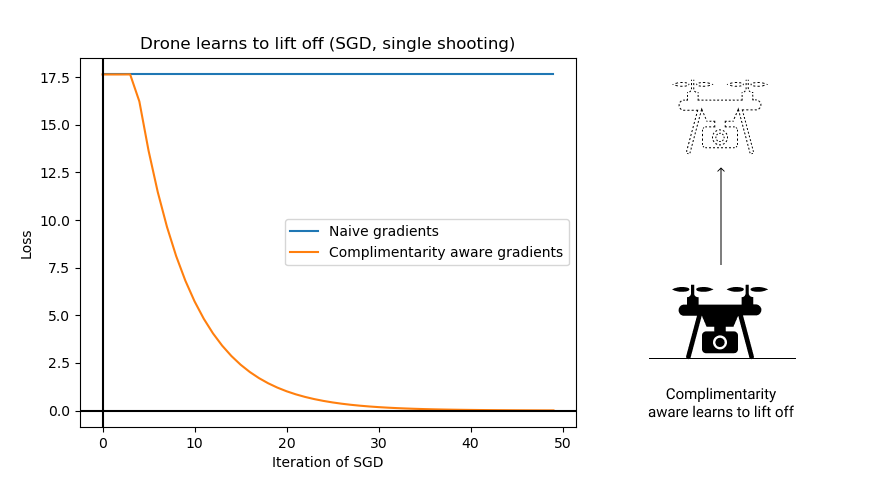}}
\caption{Training a drone to lift off the ground and fly to a target height after 500 timesteps. Loss is the squared distance of the drone from the target at $t=500$. The drone is initialized resting on the ground, which means the drone-ground contacts are classified as ``clamping.'' That means Jacobians will show $\frac{\partial \dot{\vc{q}}_{t+1}}{\partial \vc{\tau}_t} = 0$. While both standard and complementarity-aware training runs start in a saddle point, the complementarity-aware gradients are able to guide SGD to escape after several iterations of learning.}
\label{fig:complementarity-aware}
\end{center}
\end{figure}

\subsection{Optimal control with contact}

We present several trajectory optimization problems that involve complex contact dynamics, optimized using multiple shooting. We demonstrate ``Catapult'', which is a 3-dof robot that is tasked with batting a free ball towards a target, in such a way that it exactly hits the target at the desired timestep (pictured in Figure \ref{fig:gradients}). We also demonstrate ``Jump-Worm'', which is a 5-dof worm-shaped robot that is attempting to jump as high as possible at the end of the trajectory. Both of these problems involve complex contact switching throughout the course of the trajectory. We also optimize a trajectory where the ``Atlas'' robot learns to jump. See the supplementary video for the resulting trajectories.

\begin{figure}[ht]
\begin{center}
\centerline{\includegraphics[width=\columnwidth]{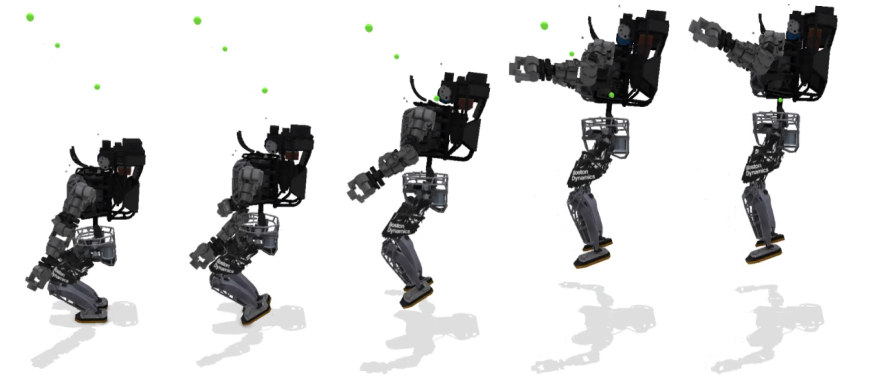}}
\caption{A few snapshots of the trajectory that an Atlas robot learns when it is asked to jump up towards the green markers, starting from a crouch.}
\label{fig:atlas}
\end{center}
\end{figure}

%% file: conclusion.tex
\section{Conclusions}

We present a fast and feature-complete differentiable physics engine for articulated rigid body simulation. We introduce a method to compute analytical gradients through the LCP formulation of inelastic contact by exploiting the sparsity of the LCP solution. Our engine supports complex contact geometry and approximating continuous-time elastic collision. We also introduce a novel method to compute complementarity-aware gradients that help optimizers to avoid stalling in saddle points.

There a few limitations of our current method. Currently, computing $\frac{\partial \vc{J}_t\vc{f}}{\partial \vc{q}_t}$, the way the joint torques produced by the contact forces change as we vary joint positions (which in turn varies contact positions and normals) takes a large portion of the total time to compute all Jacobians of dynamics for a given timestep. Perhaps a more efficient formulation can be found.

Our engine also doesn't yet differentiate through the geometric properties (like link length) of a robot, or friction coefficients. Extending to these parameters is an important step to enable parameter estimation applications.

We are excited about future work that integrates gradients \textit{and} stochastic methods to solve hard optimization problems in robotics. In running the experiments for this paper, we found that stochastic trajectory optimization methods could often find better solutions to complex problems than gradient-based methods, because they were able to escape from local optima. However, stochastic methods are notoriously sample inefficient, and have trouble fine-tuning results as they begin to approach a local optima. The authors speculate that there are many useful undiscovered techniques waiting to be invented that lie at the intersection of stochastic gradient-free methods and iterative gradient-based methods. It is our hope that an engine like the one presented in this paper will enable such research.

%% file: appendix.tex
\appendix

\begin{table*}[t]
\caption{Full Benchmarks against Finite Differencing}
\label{engine-comparison-full}
\vskip 0.1in
\begin{center}
\begin{small}
\begin{sc}
\begin{tabular}{l c c c c c c}
\toprule
Environment & Jacobian & Analytical & Central Differences  & Speedup & Ridders & Speedup  \\
 &  & Time &  Time &  & Time &  \\
\midrule
\textbf{Atlas} & All & 8.53ms & 749ms & \textbf{87.84x} & 2229ms & 261x \\
               & $\frac{\partial \bm{q}_{t+1}}{\partial \bm{q}_t}$ & 0.0204ms & 160ms & 7850x  & 581ms & 28480x\\
               & $\frac{\partial \dot{\bm{q}}_{t+1}}{\partial \bm{q}_t}$ & 6.75ms & 161ms & 23.8x  & 581ms & 86x\\
               & $\frac{\partial \bm{q}_{t+1}}{\partial \dot{\bm{q}}_t}$ & 0.016ms & 107.2ms & 6570.9x  & 304ms & 19000x\\
               & $\frac{\partial \dot{\bm{q}}_{t+1}}{\partial \dot{\bm{q}}_t}$ & 0.778ms & 160ms & 205.7x  & 487ms & 626x\\
               & $\frac{\partial \dot{\bm{q}}_{t+1}}{\partial \bm{\tau}_t}$ & 0.981ms & 161ms & 163.9x & 524ms & 471x\\
\midrule
\textbf{Half Cheetah} & All & 0.395ms & 8.64ms & \textbf{21.86x} & 24ms  & 60x \\
               & $\frac{\partial \bm{q}_{t+1}}{\partial \bm{q}_t}$ & 0.0059ms & 1.69ms & 284x  & 3.17ms & 537x \\
               & $\frac{\partial \dot{\bm{q}}_{t+1}}{\partial \bm{q}_t}$ & 0.254ms & 1.874ms & 7.35x  & 7.92ms & 31.1x \\
               & $\frac{\partial \bm{q}_{t+1}}{\partial \dot{\bm{q}}_t}$ & 0.00358ms & 1.228ms & 342x  & 2.42ms & 675.9x \\
               & $\frac{\partial \dot{\bm{q}}_{t+1}}{\partial \dot{\bm{q}}_t}$ & 0.053ms & 1.98ms & 37.1x & 5.63ms & 106.2x \\
               & $\frac{\partial \dot{\bm{q}}_{t+1}}{\partial \bm{\tau}_t}$ & 0.077ms & 1.87ms & 24.1x & 5.75ms & 74.6x \\
\midrule
\textbf{Jump-Worm} & All & 0.256ms & 3.82ms & \textbf{14.89x} & 9.727ms & 37.9x \\
               & $\frac{\partial \bm{q}_{t+1}}{\partial \bm{q}_t}$ & 0.00433ms & 0.732ms & 168.8x & 1.58ms & 364.8x \\
               & $\frac{\partial \dot{\bm{q}}_{t+1}}{\partial \bm{q}_t}$ & 0.174ms & 0.836ms & 4.78x  & 2.84ms & 16.32x \\
               & $\frac{\partial \bm{q}_{t+1}}{\partial \dot{\bm{q}}_t}$ & 0.00307ms & 0.532ms & 173.1x  & 0.81ms & 263.8x \\
               & $\frac{\partial \dot{\bm{q}}_{t+1}}{\partial \dot{\bm{q}}_t}$ & 0.0343ms & 0.872ms & 25.4x  & 2.17ms & 63.2x \\
               & $\frac{\partial \dot{\bm{q}}_{t+1}}{\partial \bm{\tau}_t}$ & 0.040ms & 0.851ms & 21.2x & 2.31ms & 57.75x \\
\midrule
\textbf{Catapult} & All & 0.265ms & 4.36ms & \textbf{16.45x} & 12.0ms & 45.2x \\
               & $\frac{\partial \bm{q}_{t+1}}{\partial \bm{q}_t}$ & 0.0050ms & 0.845ms & 167.8x & 1.79ms & 358x \\
               & $\frac{\partial \dot{\bm{q}}_{t+1}}{\partial \bm{q}_t}$ & 0.181ms & 0.972ms & 5.36x  & 3.71ms & 20.49x \\
               & $\frac{\partial \bm{q}_{t+1}}{\partial \dot{\bm{q}}_t}$ & 0.0036ms & 0.574ms & 156.7x  & 0.958ms & 266x \\
               & $\frac{\partial \dot{\bm{q}}_{t+1}}{\partial \dot{\bm{q}}_t}$ & 0.035ms & 1.01ms & 28.42x  & 2.72ms & 77.7x \\
               & $\frac{\partial \dot{\bm{q}}_{t+1}}{\partial \bm{\tau}_t}$ & 0.039ms & 0.960ms & 24.3x & 2.85ms & 73.1x \\
               
\bottomrule
\end{tabular}
\end{sc}
\end{small}
\end{center}
\vskip -0.1in
\end{table*}

\subsection{Frictional impulse}
\label{appendix:friction}

In DART, friction impulses are solved by the boxed LCP method, using the same implementation of the boxed variant of the Dantzig algorithm found in the Open Dynamics Engine, and originally proposed in \cite{baraff1994fast}. Boxed LCPs are not theoretically guaranteed to be solvable, but are quite common in practice because of their speed and high-quality results. We therefore extend our formulation in Section \ref{sec:lcp} to compute the gradient of frictional impulse magnitudes found in a boxed LCP solver with respect to $\vc{A}$ and $\vc{b}$. Similar to normal impulses, each frictional impulse is classified into one of the two states: 

\paragraph{Clamping ($\mathcal{C}$)} If the relative velocity along the frictional impulse direction is zero, and friction impulse magnitude is below its bound, then any attempt to push this contact will be met with an increase in frictional impulse holding the point in place. This means the contact point is clamping, which behaves will be treated in the same way as clamped normal forces.

\paragraph{Bounded ($\mathcal{B}$)} If the frictional impulse magnitude is at its bound (either positive or negative) then the contact point is sliding or about to slide along this friction direction. Bounded frictional impulse $\mathcal{B}$ is quite like ``Separating'' $\mathcal{S}$ in normal impulse. The difference is that frictional impulses in $\mathcal{B}$ are not zero but at a non-zero bound based on the corresponding normal impulses.

Bounded frictional impulse can be expressed by $\vc{f}^*_{\mathcal{B}} = \vc{E}\vc{f}^*_{\mathcal{C}}$, where each row of $\vc{E} \in \mathcal{R}^{|\mathcal{B}| \times |\mathcal{C}|}$ contains a single non-zero value with the friction coefficient, $\mu_s$, for the corresponding normal impulse index in $\mathcal{C}$.

We define $\vc{J}_{\mathcal{C}}$ to be the matrix with just the columns of $\vc{J}$ corresponding to the indices in $\mathcal{C}$. Likewise, we define $\vc{J}_{\mathcal{B}}$ containing just the columns of $\vc{J}$ that are bounded. If we multiply $\vc{J}^T_{\mathcal{C}}\vc{f}_{\mathcal{C}}$ we get the joint torques due to the clamping constraint forces. Similarly, if we multiply $\vc{J}^T_{\mathcal{B}}\vc{f}_{\mathcal{B}}$ we get the joint torques due to bounded friction impulses. Since $\vc{f}_{\mathcal{B}} = \vc{E}\vc{f}_{\mathcal{C}}$, we can modify $\vc{A}_{\mathcal{CC}}$ to take bounded frictional impulses into account:
\[\vc{A}_{\mathcal{CC}} = \vc{J}_{\mathcal{C}} \vc{M}^{-1}(\vc{J}^T_{\mathcal{C}} + \vc{J}^T_{\mathcal{B}}\vc{E})\]

% \[A_{\mathcal{CC}} = J_{\mathcal{C}}^TM^{-1}J_{\mathcal{C}}\]

% In the above, we take $J_{\mathcal{C}}$ to be the matrix with just the columns of $J$ corresponding to $\mathcal{C}$. Likewise, we now construct $J_{\mathcal{U}}$, containing just the columns of $J$ that are upper bounded. If we multiply $J_{\mathcal{C}}f_{\mathcal{C}}$ we get the joint torques due to the clamping constraint forces. Similarly, if we multiply $J_{\mathcal{U}}f_{\mathcal{U}}$ we get the joint torques due to the upper bounded constraint forces. Since we have $f_{\mathcal{U}} = Ef_{\mathcal{C}}$, we could also multiply $J_{\mathcal{U}}Ef_{\mathcal{C}}$ to get the joint torques due to upper bounded constraint forces. With that intuition in mind, we can make a small modification to $A_{\mathcal{CC}}$ to take upper bounded constraints into account:

% \[A_{\mathcal{CC}} = J_{\mathcal{C}}^TM^{-1}(J_{\mathcal{C}} + J_{\mathcal{U}}E)\]

With this one small change, all the formulation in Section \ref{sec:lcp} works with frictional forces. An interesting observation is that the bounded frictional cases are analogous to the separating cases for the normal forces, in that the force value is constrained at $\mu_s f_i$ (or $-\mu_s f_i$), where $f_i$ is the corresponding normal force for the same contact. The only way the bounded force values will change is through the change of corresponding clamping normal force, which needs to be accounted for when computing the gradients.

Just like the overall boxed LCP problem is not solvable, $\vc{A}_{\mathcal{CC}}$ is no longer guaranteed to be exactly invertible. In order to support this, we need to use the pseudoinverse of $\vc{A}_{\mathcal{CC}}$ during the forward pass, and the gradient of the pseudoinverse when computing Jacobians.

\subsection{LCP stabilization}
\label{appendix:stabilize}

When $\vc{A}_{\mathcal{CC}}$ is not full rank, the solution to $f_{\text{LCP}}(\vc{A}, \vc{b}) = \vc{f}^{*}$ is no longer unique. To grasp this intuitively, consider a 2D case where a box of unit mass that cannot rotate is resting on a plane. The box has two contact points, with identical contact normals, and because the box is not allowed to rotate, the effect of an impulse at each contact point is exactly the same (it causes the box's upward velocity to increase). This means that both columns of $\vc{A}$ (one per contact) are identical. That means that $\vc{A}_{\mathcal{CC}} \in \mathcal{R}^{2 \times 2}$ is actually only rank one. Let's assume we need a total upward impulse of $-mg$ to prevent the box from interpenetrating the floor. Because $\vc{A}_{\mathcal{CC}}$ is a low rank, we're left with one equation and two unknowns:

\[\vc{A}_{\mathcal{CC}}\vc{f}^{*}_{\mathcal{C}} = \begin{bmatrix}
1 && 1 \\
1 && 1
\end{bmatrix}\begin{bmatrix}f^{*}_{{\mathcal{C}_1}} \\ f^{*}_{{\mathcal{C}}_2}\end{bmatrix} = \begin{bmatrix} f^{*}_{{\mathcal{C}}_1} + f^{*}_{{\mathcal{C}}_2} \\ f^{*}_{{\mathcal{C}}_1} + f^{*}_{{\mathcal{C}}_2} \end{bmatrix} = \begin{bmatrix} -mg \\ -mg \end{bmatrix}\]

It is easy to see that this is just $f^{*}_{{\mathcal{C}}_1} + f^{*}_{{\mathcal{C}}_2} = -mg$. And that means that we have an infinite number of valid solutions to the LCP.

In order to have valid gradients, our LCP needs to have predictable behavior when faced with multiple valid solutions. Thankfully, our analysis in the previous sections suggests a quite simple and efficient (and to the authors' knowledge novel) LCP stabilization method. Once an initial solution is computed using any algorithm, and the clamping set $\mathcal{C}$ is found, we can produce a least-squares-minimal (and numerically exact) solution to the LCP by setting:

\[\vc{f}^{*}_{\mathcal{C}} = \vc{A}_{\mathcal{CC}}^{-1}\vc{b}_{\mathcal{C}}, \hspace{3em} \vc{f}^{*}_{\mathcal{S}} = 0\]

This is possible with a single matrix inversion because the hard part of the LCP problem (determining which indices belong in which classes) was already solved for us by the main solver. Once we know which indices belong in which classes, solving the LCP exactly reduces to simple linear algebra.

As an interesting aside, this ``LCP stabilization'' method doubles as an extremely efficient LCP solver for iterative LCP problems. In practical physics engines, most contact points do not change from clamping ($\mathcal{C}$) to separating ($\mathcal{S}$) or back again on most time steps. With that intuition in mind, we can opportunistically attempt to solve a new $f_{\text{LCP}}(\vc{A}_{t+1}, \vc{b}_{t+1}) = \vc{f}_{t+1}^{*}$ at a new timestep by simply guessing that the contacts will sort into $\mathcal{C}$ and $\mathcal{S}$ in exactly the same way they did on the last time step. Then we can solve our stabilization equations for $\vc{f}_{t+1}^{*}$ as follows:

\[{\vc{f}_{t+1}}^{*}_{\mathcal{C}} = {\vc{A}_{t+1}}_{\mathcal{CC}}^{-1}{\vc{b}_{t+1}}_{\mathcal{C}}, \hspace{3em} {\vc{f}_{t+1}}^{*}_{\mathcal{S}} = 0\]

If we guessed correctly, which we can verify in negligible time, then $\vc{f}_{t+1}^{*}$ is a valid, stable, \textit{and perfectly numerically exact} solution to the LCP. When that happens, and in our experiments this heuristic is right $> 95\%$ of the time, we can skip the expensive call to our LCP solver entirely. As an added bonus, because $\mathcal{C}$ is usually not all indices, inverting ${\vc{A}_{t+1}}_{\mathcal{CC}}$ can be considerably cheaper than inverting all of $\vc{A}_{t+1}$, which can be necessary in an algorithm to solve the full LCP.

When our heuristic does not result in a valid $\vc{f}_{t+1}^{*}$, we can simply return to our ordinary LCP solver to get a valid set $\mathcal{C}$ and $\mathcal{S}$, and then re-run our stabilization.

As long as results from our LCP are stabilized, the gradients through the LCP presented in this section are valid even when $\vc{A}$ is not full rank.

\subsection{Contact point and normal stabilization and gradients}

\label{appendix:contact}

To compute gradients through each contact point and normal with respect to $\vc{q}_t$, we need to provide specialized routines for each type of collision geometry, including collisions between spheres, capsules, boxes, and arbitrary convex meshes. Deriving the routines is a matter of straightforward algebra. However in order for well-defined gradients to exist at all, each collision must have precisely specified contact points and normals, so that we can compute well-behaved gradients.

We describe in this appendix how we produce well-defined contact points and normals, since this is a design choice. The gradients of these methods are left as an exercise to the reader, who is invited to check their work against our open-source implementation.

\subsubsection{Mesh-mesh collisions}

Mesh-mesh collisions only have two types of contacts: vertex-face and edge-edge collisions. The other cases (vertex-edge, vertex-vertex, edge-face, face-face) are degenerate and easily mapped into vertex-face and edge-edge collisions.

Mesh-mesh collision detection algorithms like Gilbert-Johnson-Keerthi or Minkowski Portal Refinement are iterative, and so produce imprecise collision points and normals that can vary depending on initialization. Both algorithms produce a separating ``witness plane'' as part of their output, which is a 3D plane that approximately separates the two meshes (as much as possible, if they're intersecting). Going from an approximate separating witness plane to precisely specified vertex-face and edge-edge collisions in the general case is complex. Take all the points on object A that lie within some tiny $\epsilon$ of the witness plane, and map them into 2D coordinates within the witness plane. Do likewise with the points on object B. Now we have two convex 2D shapes, because any linear slice of a convex shape is itself convex. Throw out any vertices that are not on the 2D convex hull of the point cloud for A and B respectively. Call the resulting convex shapes ``witness hulls'' of A and B. Now any vertices on the witness hull of A that lie within the witness hull of B are vertex-face collisions from A to B. Analogously, any vertices on the witness hull of B that lie within the witness hull of A are face-vertex collisions from A to B. Finally, any edges of the witness hulls A and B that intersect are edge-edge collisions.

For all vertex-face collisions, during the forward simulation, the collision detector places a collision at the point of the vertex, with a normal dictated by the face under collision. The body providing the vertex can only influence the collision location $\vc{p}$, and the body providing the face can only influence the collision normal $\vc{n}$.

For all edge-edge collisions, the collision detector first finds the nearest point on edge A to edge B (call that $\vc{a}$), and the nearest point on edge B to edge A (call that $\vc{b}$). Then the contact point is set to the average of $\vc{a}$ and $\vc{b}$, which is $\frac{\vc{a} + \vc{b}}{2}$. The normal is given by the cross product of the two edges. That means changing $\vc{q}$ of Object A can affect the contact normal and the contact location along the other edge from Object B. For this reason, we need to construct our Jacobians globally.

\subsubsection{Sphere-sphere collisions}

These are straightforward. We have sphere A, with center $c_a$ and radius $r_a$, and sphere B, with center $c_b$ and radius $r_b$. The contact normal is the normalized vector pointing from $c_b$ to $c_a$. The contact point is a weighted combination of the two sphere centers, $\frac{r_b*c_a + r_a*c_b}{r_a + r_b}$.

\subsubsection{Mesh-sphere collisions}

These can be divided into three categories: sphere-face collisions, sphere-edge collisions, and sphere-vertex collisions. Starting from the simplest, a sphere-vertex collision places the contact point at the vertex and the normal points from the vertex to the sphere center.

A sphere-edge collision places the contact point on the closest point to the sphere center along the edge. The contact normal points from the collision point to the sphere center.

A sphere-face collision gets the contact normal from the normal of the colliding face. Call the contact normal $\vc{n}$, and define the sphere center as $\vc{c}$ and radius as $\vc{r}$. For the simplicity of downstream gradient computation, we then place the contact point at the point on the sphere, projected along with the contact normal towards the contact: $\vc{c} + \vc{r}*\vc{n}$.

\subsubsection{Pipe-pipe collisions}

In a capsule-capsule collision, when both capsule cylinders are colliding, we call that a pipe-pipe collision. Mechanically, this is very similar to an edge-edge collision, only with added radius variables $r_a$ and $r_b$. Let $c_a$ be the nearest point on the centerline of pipe A to the centerline of pipe B. Let $c_b$ be the nearest point on the centerline of pipe B to the centerline of pipe A. Then we say the contact point is $\frac{r_b*c_a + r_a*c_b}{r_a + r_b}$. The contact normal points from $c_b$ to $c_a$.

\subsubsection{Pipe-sphere collisions}

These look a lot like pipe-pipe collisions, except that $c_b$ is now fixed to the center of the sphere. Let $c_a$ be the nearest point on the centerline of the pipe to $c_b$. Then we say the contact point is $\frac{r_b*c_a + r_a*c_b}{r_a + r_b}$. The contact normal points from $c_b$ to $c_a$.

\subsubsection{Pipe-mesh collisions}

This breaks down into two types of contact: vertex-pipe, and edge-pipe. Face-pipe contacts reduce to two edge-pipe contacts.

For vertex-pipe contacts, we put the contact point at the vertex, and point the normal towards the nearest point on the centerline of the pipe.

For edge-pipe contacts, we treat them as pipe-pipe contacts where the radius of the first pipe is 0.

\subsubsection{Gradients}

Once all the contact behavior is firmly established, it's just a matter of rote calculus to compute derivatives. We refer the reader to our open source code for the implementation of those derivatives. Once a stable collision detection system and its derivatives are implemented that, it's possible to efficiently compute $\frac{\partial J_t^Tf}{\partial q_t}$.

\subsection{Jacobian Benchmark Evaluation}

\label{appendix:benchmarks}

In addition to speed, we are interested in the accuracy of our Jacobians, as gradients computed via finite differencing become more inaccurate as the system becomes more complex, which can lead to instability in optimization. As another baseline, we apply Ridders' method \cite{ridders1982} to efficiently calculate Jacobians to a higher-order error than central differencing, using the stopping criterion given in \cite{numericalrecipes}. 
% We show that our analytical Jacobians are significantly more accurate than those computed by central differencing, while also being much faster to evaluate.

Table \ref{engine-comparison-full} contains the full benchmark performance of our analytical Jacobians against central differencing and the accurate Ridders' extrapolated finite differences. For each environment, we compare the speed of evaluation of each individual component Jacobian as well as the total time. %\karen{Can we also show the time and speed up against Ridder?} \dalton{We can, I took it out because it doesn't make too much sense to benchmark time against something no one would realistically use, and because it adds 2 columns to the table. but I could put it back}. We compare accuracy by the $L_\infty$ difference between our analytical solutions and the Ridders baseline solution, as well as the difference between central differencing and Ridders. Tests in Table \ref{engine-comparison-full} were computed on an AMD® Ryzen 5 2600x processor. \dalton{The vel-pos Atlas jacobian is inaccurate, likely due to a bug in the jacobian. One jacobian noted in Catapult is also SUPER bad, for some reason.}

\subsection{Calculating the Bounce Approximation Jacobian}

\label{appendix:bounces}

Recall the matrix $\vc{J}^T$ that transforms contact forces to joint forces. By conservation of momentum $\vc{J}\dot{\vc{q}}_t = \vc{v}_t$, leading to:

\begin{equation}
\frac{\partial \vc{p}_{t+1}}{\partial \vc{p}_{t}} \approx \vc{J} \frac{\partial \vc{q}_{t+1}}{\partial \vc{q}_t} \vc{J}^{-1}
\end{equation}

Our matrix notation is somewhat misleading here, because we do not want our approximation to capture off-diagonals of $\frac{\partial \vc{p}_{t+1}}{\partial \vc{p}_{t}}$. Because we construct our approximate $\frac{\partial \vc{p}_{t+1}}{\partial \vc{p}_{t}}$ by assuming each bounce is independent, we end up with 0s on the off-diagonals, but we know that assumption to be false. We just want to find a $\frac{\partial \vc{q}_{t+1}}{\partial \vc{q}_t}$ where the above equation matches the diagonals as closely as possible, ignoring other elements. The following derives this relation in closed form.

Since we are only interested in enforcing the diagonal entries, so we can write out a series of linear constraints we would like to attempt to satisfy. Let $\vc{J}_{(i)}$ denote the $i$'th column of $\vc{J}$. Recall that $\frac{\partial \vc{p}_{i,t+1}}{\partial \vc{p}_{i,t}} = -\sigma_i$.

\begin{equation}
\vc{J}_{(i)} \frac{\partial \vc{q}_{t+1}}{\partial \vc{q}_t} \vc{J}^{-1}_{(i)} \approx -\sigma_i
\end{equation}

We would like to find some approximate matrix $\frac{\partial \vc{q}_{t+1}}{\partial \vc{q}_t}$ that satisfies all $n$ of the above constraints as closely as possible. Stated directly as a least-squares optimization object:

\begin{equation}
\text{min} \sum_i \big( \vc{J}_{(i)} \frac{\partial \vc{q}_{t+1}}{\partial \vc{q}_t} \vc{J}^{-1}_{(i)} - (-\sigma_i) \big)^2
\end{equation}

The solution to this optimization can be found without iterative methods.

Let $\frac{\partial \vc{q}_{t+1}}{\partial \vc{q}_t}_{(i)}$ be the $i$'th column of $\frac{\partial \vc{q}_{t+1}}{\partial \vc{q}_t}$, and $\vc{J}_{(j,i)}$ be the $j$'th row and the $i$'th column of $\vc{J}$.
Note that:
\begin{equation}
\vc{J}_{(i)} \frac{\partial \vc{q}_{t+1}}{\partial \vc{q}_t} \vc{J}^{-1}_{(i)} = \sum_j \underbrace{\vc{J}_{(j,i)}}_{\text{scalar}}(\vc{J}^{-T}_{(i)} \frac{\partial \vc{q}_{t+1}}{\partial \vc{q}_t}_{(j)})
\end{equation}

\begin{equation}
\underbrace{\vc{J}_{(j,i)}}_{\text{scalar}}(\vc{J}^{-T}_{(i)} \frac{\partial \vc{q}_{t+1}}{\partial \vc{q}_t}_{(j)}) = \sum_j {\underbrace{(\vc{J}_{(j,i)}\vc{J}^{-T}_{(i)})}_{\text{vector}}}^T\frac{\partial \vc{q}_{t+1}}{\partial \vc{q}_t}_{(j)})
\end{equation}

It becomes clear that we could construct a long vector $\vc{v}$, which will map to every column of $\frac{\partial \vc{q}_{t+1}}{\partial \vc{q}_t}$ placed end to end. We can also construct a matrix $\vc{W}$ where every column $\vc{W}_{(i)}$ is the vectors $\vc{J}_{(j,i)}\vc{J}^{-1}_{(i)}$ placed end to end (iterating over $j$).
Now if we take the values of $\sigma_{i}$ as entries of a vector $\vc{r} \in \mathcal{R}^n$, we can write our optimization problem as a linear equation:

\begin{equation}
\text{min} \sum_i \big( \vc{J}_{(i)} \frac{\partial \vc{q}_{t+1}}{\partial \vc{q}_t} \vc{J}^{-1}_{(i)} - (-\sigma_i) \big)^2
 = \text{min } || \vc{W}^T\vc{v} + \vc{r} ||_2^2
\end{equation}
 
This is a standard least squares problem, and is solved when:

\begin{equation}
\vc{v} = -\vc{W}^{T \dagger}\vc{r}
\end{equation}

Once we have a value of $\vc{v}$, we can reconstruct the original matrix $\frac{\partial \vc{q}_{t+1}}{\partial \vc{q}_t}$ by taking each column of $\frac{\partial \vc{q}_{t+1}}{\partial \vc{q}_t}$ the appropriate segment of $\vc{v}$.

This is almost always an under-determined system, and we want to default to having $\frac{\partial \vc{q}_{t+1}}{\partial \vc{q}_t}$ as close to $I$ as possible, rather than as close to 0 as possible. We can slightly reformulate our optimization problem where the least square objective tries to keep the diagonals at 1, rather than 0. If we define an arbitrary $\vc{c}$ vector (for ``center''), we can use the identity:

\begin{equation}
\vc{W}^T(\vc{v} - \vc{c}) = -\vc{r} - \vc{W}^T\vc{c}
\end{equation}
\begin{equation}
\vc{v} = \vc{c} - \vc{W}^{T\dagger}(\vc{r} + \vc{W}^T\vc{c})
\end{equation}

If we set $\vc{c}$ to the mapping for $\frac{\partial \vc{q}_{t+1}}{\partial \vc{q}_t} = I$, then we get a solution that minimizes the distance to the identity while satisfying the constraints, measured as the sum of the squares of all the terms.

Also, remember that to get our Jacobian with respect to velocity, we simply:

\begin{equation}
\frac{d\theta_{t+1}}{d\dot{\theta}_t} = \Delta t \frac{\partial \vc{q}_{t+1}}{\partial \vc{q}_t}
\end{equation}

And that should approximately solve the ``gradient bounce'' problem. On timesteps where there is only a single bouncing contact, this will provide an exact solution. With more than one bounce in a single frame, this may be approximate.

\subsection{Analytical derivatives through Featherstone}
\label{appendix:featherstone}
We present the derivation of analytical derivatives computation through the Featherstone algorithm. Although the intellectual computation should be credited entirely to \cite{kim2012lie} and \cite{carpentier2018analytical}, we specialized the derivations to obtain $\frac{\partial \vc{M}_t^{-1} \vc{z}_t}{\partial \vc{q}_t}$, $\frac{\partial \vc{c}_t}{\partial \vc{q}_t}$, and $\frac{\partial \vc{c}_t}{\partial \dot{\vc{q}}_t}$ for our implementation, rather than the entire inverse and forward dynamics. The detailed derivation might be of interest to some readers.
% \karen{JS, please complete this paragraph.}
% \js{Keenon, do we want to describe the entire derivative of inverse dynamics and forward dynamics, or just specific components that are $\pd{(C\dot{q} + g)}{q}, $\pd{(C\dot{q} + g)}{\dot{q}}, and $\pd{M^{-1}}{q}$?}

% $\pd{h_i}{q_j} = \begin{cases} S_i^k & \text{if } q_j = q_i^k \\ 0, & \text{otherwise} \end{cases}$, $\pd{S_i}{q_j} = \pd{\dot{S}_i}{q_j} = 0$ when $q_j \centernot\in q_i$.

The partial derivative of the inverse of joint space inertia matrix can be computed from the partial derivative of the joint space inertia matrix through the relation \cite{carpentier2018analytical}:
\begin{align}
\pd{\vc{M}^{-1} \vc{z}}{\vc{q}} &= \vc{M}^{-1} \pd{\vc{M}}{\vc{q}} \vc{M}^{-1} \vc{z}.
\end{align}
Algorithm \ref{alg:fd_for_minv} and \ref{alg:id_deriv_for_m} show the recursive algorithms to compute $\vc{M}^{-1}$ and $\pd{\vc{M}}{\vc{q}} \vc{M}^{-1} \vc{z}$, respectively. In Algorithm \ref{alg:id_deriv_for_c}, the derivatives of the Coriolis force with respect to the joint position and the joint velocity are given.

\begin{algorithm}[H]
	\caption{Recursive forward dynamics for $\vc{M}^{-1}$}
	\label{alg:fd_for_minv}
    \begin{algorithmic}[1]
	\For{$j = 1 \text{ to } n$}
        \For{$i = n \text{ to } 1$}
            \State{$\AI_i = \I_i + \sum_{l \in \mu(i)} \dAd{\vc{T}_{i,l}^{-1}} \vc{\Pi}_l \Ad{\vc{T}_{i,l}^{-1}}$}
            \State{$\AB_i = \sum_{l \in \mu(i)} \dAd{\vc{T}_{i,l}^{-1}} \vc{\beta}_l$}
            \State{$\vc{\Psi}_i = \left( \vc{S}_i^T \AI_i \vc{S}_i \right)^{-1}$}
            \State{$\vc{\Pi}_i = \AI_i - \AI_i \vc{S}_i \Psi_i \vc{S}_i^T \AI_i$}
            \State{$\vc{\alpha}_i = \delta_{i, j} - \vc{S}_i^T \AB_i$, where $\delta_{i,j} = \begin{cases} 1 & \text{if } i = j \\ 0, & \text{otherwise} \end{cases}$}
            \State{$\vc{\beta}_i = \AB_i + \AI_i \vc{S}_i \vc{\Psi}_i \vc{\alpha}_i$}
        \EndFor
        \For{$i = 1 \text{ to } n$}
            \State{$\left[ \vc{M}^{-1} \right]_{i, j} = \vc{\Psi}_i \left( \vc{\alpha}_i - \vc{S}_i^T \AI_i \AdTi \dot{\vc{V}}_{\lambda(i)} \right)$}
            \State{$\dot{\vc{V}}_i = \AdTi \dot{\vc{V}}_{\lambda(i)} + \vc{S}_i \left[ \vc{M}^{-1} \right]_{i, j} $}
        \EndFor
    \EndFor
    \end{algorithmic}
\end{algorithm}

\begin{algorithm}[H]
	\caption{Derivative of the recursive inverse dynamics for $\pd{\vc{M}}{\vc{q}} \vc{M}^{-1} \vc{z}$}
	\label{alg:id_deriv_for_m}
    \begin{algorithmic}[1]
    \For{$i = 1 \text{ to } n$}
        \State{$\dot{\vc{V}}_i = \AdTi \dot{\vc{V}}_{\lambda(i)} + \vc{S}_i \left[ \vc{M}^{-1} \vc{z} \right]_i$}
        \For{$j = 1 \text{ to } n$}
            \State{$\pd{\dot{\vc{V}}_i}{\vc{q}_j} = \AdTi \pd{\dot{\vc{V}}_{\lambda(i)}}{\vc{q}_j} - \ad{\pd{\vc{h}_i}{\vc{q}_j}} \AdTi \dot{\vc{V}}_{\lambda(i)}$}
            \State{$~~~~~~~~ + \pd{\vc{S}_i}{\vc{q}_j} \left[ \vc{M}^{-1} \vc{z} \right]_i$}
        \EndFor
    \EndFor
    \For{$i = n \text{ to } 1$}
        \State{$\vc{F}_i = \I_i \dot{\vc{V}}_i + \sum_{l \in \mu(i)} \dAd{\vc{T}_{i,l}^{-1}} \vc{F}_l$}
        \For{$j = 1 \text{ to } n$}
            \State{$\pd{\vc{F}_i}{\vc{q}_j} = \I_i \pd{\dot{\vc{V}}_i}{\vc{q}_j} - \pd{\vc{F}^{\text{g}}}{\vc{q}_j}$}
            \State{$~~~~~~~~~ + \sum_{l \in \mu(i)} \dAd{\vc{T}_{i,l}^{-1}} \left( \pd{\vc{F}_l}{\vc{q}_j} - \dad{\pd{\vc{h}_i}{\vc{q}_j}} \vc{F}_l \right)$}
            \State{$\left[ \pd{\vc{M}}{\vc{q}} \vc{M}^{-1} \vc{z} \right]_{i,j} = \pd{\vc{S}_i}{\vc{q}_j}^T \vc{F}_i + \vc{S}_i^T \pd{\vc{F}_i}{\vc{q}_j}$}
        \EndFor
    \EndFor
    \end{algorithmic}
\end{algorithm}

\begin{algorithm}[H]
	\caption{Derivative of the recursive inverse dynamics for $\pd{\vc{c}}{\vc{q}}$ and $\pd{\vc{c}}{\dot{\vc{q}}}$}
	\label{alg:id_deriv_for_c}
    \begin{algorithmic}[1]
    \For{$i = 1 \text{ to } n$}
        \State{$\vc{V}_i = \AdTi \vc{V}_{\lambda(i)} + \vc{S}_i \dot{\vc{q}}_i$}
        \State{$\dot{\vc{V}}_i = \AdTi \dot{\vc{V}}_{\lambda(i)} + \ad{\vc{V}_i} \vc{S}_i \dot{\vc{q}}_i + \dot{\vc{S}}_i \dot{\vc{q}}_i$}
        \For{$j = 1 \text{ to } n$}
            \State{$\pd{\vc{V}_i}{\vc{q}_j} = \AdTi \pd{\vc{V}_{\lambda(i)}}{\vc{q}_j} - \ad{\pd{\vc{h}_i}{\vc{q}_j}} \AdTi \vc{V}_{\lambda(i)} + \pd{\vc{S}_i}{\vc{q}_j} \dot{\vc{q}}_i$}
            \State{$\pd{\vc{V}_i}{\dot{\vc{q}}_j} = \AdTi \pd{\vc{V}_{\lambda(i)}}{\dot{\vc{q}}_j} + \vc{S}_i^k$}
            \State{$\pd{\dot{\vc{V}}_i}{\vc{q}_j} = \AdTi \pd{\dot{\vc{V}}_{\lambda(i)}}{\vc{q}_j} - \ad{\pd{\vc{h}_i}{\vc{q}_j}} \AdTi \dot{\vc{V}}_{\lambda(i)}$}
            \State{$~~~~~~~~ + \ad{\pd{\vc{V}_i}{\vc{q}_j}} \vc{S}_i \dot{\vc{q}}_i + \ad{\vc{V}_i} \pd{\vc{S}_i}{\vc{q}_j} \dot{\vc{q}}_i + \pd{\dot{\vc{S}}_i}{\vc{q}_j} \dot{\vc{q}}_i$}
            \State{$\pd{\dot{\vc{V}}_i}{\dot{\vc{q}}_j} = \AdTi \pd{\dot{\vc{V}}_{\lambda(i)}}{\dot{\vc{q}}_j}$}
            \State{$~~~~~~~~ + \ad{\pd{\vc{V}_i}{\dot{\vc{q}}_j}} \vc{S}_i \dot{\vc{q}}_i + \ad{\vc{V}_i} \vc{S}_i^k + \pd{\dot{\vc{S}}_i}{\dot{\vc{q}}_j} \dot{\vc{q}}_i + \dot{\vc{S}}_i^k$}
        \EndFor
    \EndFor
    \For{$i = n \text{ to } 1$}
        \State{$\vc{F}_i = \I_i \dot{\vc{V}}_i - \dad{\vc{V}_i} \I_i \vc{V}_i - \vc{F}_i^g + \sum_{l \in \mu(i)} \dAd{\vc{T}_{i,l}^{-1}} \vc{F}_l$}
        \For{$j = 1 \text{ to } n$}
            \State{$\pd{\vc{F}_i}{\vc{q}_j} = \I_i \pd{\dot{\vc{V}}_i}{q_j} - \dad{\pd{\vc{V}_i}{q_j}} \I \vc{V}_i - \dad{\vc{V}_i} \I_i \pd{\vc{V}_i}{\vc{q}_j} - \pd{\vc{F}^{\text{g}}}{\vc{q}_j}$}
            \State{$~~~~~~~~~ + \sum_{l \in \mu(i)} \dAd{\vc{T}_{i,l}^{-1}} \left( \pd{\vc{F}_l}{\vc{q}_j} - \dad{\pd{\vc{h}_i}{\vc{q}_j}} \vc{F}_l \right)$}
            \State{$\pd{\vc{F}_i}{\dot{\vc{q}}_j} = \I_i \pd{\dot{\vc{V}}_i}{\dot{\vc{q}}_j} - \dad{\pd{\vc{V}_i}{\dot{\vc{q}}_j}} \I \vc{V}_i - \dad{\vc{V}_i} \I_i \pd{\vc{V}_i}{\dot{\vc{q}}_j}$}
            \State{$~~~~~~~~~ + \sum_{l \in \mu(i)} \dAd{\vc{T}_{i,l}^{-1}} \pd{\vc{F}_l}{\vc{q}_j}$}
            \State{$\left[ \pd{\vc{c}}{\vc{q}} \right]_{i,j} = \pd{\vc{S}_i}{\vc{q}_j}^T \vc{F}_i + \vc{S}_i^T \pd{\vc{F}_i}{\vc{q}_j}$}
            \State{$\left[ \pd{\vc{c}}{\dot{\vc{q}}} \right]_{i,j} = \vc{S}_i^T \pd{\vc{F}_i}{\dot{\vc{q}}_j}$}
        \EndFor
    \EndFor
    \end{algorithmic}
\end{algorithm}